\documentclass[journal]{IEEEtran}

\usepackage{graphicx}
\usepackage{amsmath}
\usepackage{amsfonts}
\usepackage{multicol}
\usepackage{multirow}
\usepackage{cite}
\usepackage{amsmath}
\usepackage{color,xcolor}
\usepackage{color}
\usepackage[pagebackref=true,breaklinks=true,letterpaper=true,colorlinks,bookmarks=false]{hyperref}

\newcommand{\etal}{\textit{et al}.}
\newcommand{\ie}{\textit{i}.\textit{e}.}
\newcommand{\eg}{\textit{e}.\textit{g}.}
\newcommand{\etc}{\textit{etc}}

\begin{document}

\title{An Underwater Image Enhancement Benchmark Dataset and Beyond}
\author{Chongyi~Li,
        Chunle~Guo,
        Wenqi~Ren,~\IEEEmembership{Member,~IEEE,}
        Runmin Cong,~\IEEEmembership{Member,~IEEE,}
        Junhui~Hou,~\IEEEmembership{Member,~IEEE,} \\
        Sam Kwong,~\IEEEmembership{Fellow,~IEEE,}
        and Dacheng Tao,~\IEEEmembership{Fellow,~IEEE}
\thanks{This work was supported in part by CCF-Tencent Open Fund, in part by Zhejiang Lab's International Talent Fund for Young Professionals, in part by Fundamental Research Funds for the Central Universities under Grant 2019RC039, in part by National Natural Science Foundation of China under Grant 61802403, Grant 61771334, and Grant 61871342,  in part by Hong Kong RGC General Research Funds under Grant 9042038 (CityU 11205314) and Grant 9042322 (CityU 11200116), and in part by Hong Kong RGC Early Career Schemes under Grant 9048123. (\emph{Corresponding author: Wenqi Ren})}
\thanks{Chongyi Li is with the Department of Computer Science, City University of Hong Kong, Kowloon 999077, Hong Kong SAR, China, and also with State Key Laboratory of Information Security, Institute of Information Engineering, Chinese Academy of Sciences, China  (e-mail: lichongyi25@gmail.com).

Chunle Guo is with the School of Electrical and Information Engineering, Tianjin University, Tianjin, China (e-mail: guochunle@tju.edu.cn).

Wenqi Ren is with State Key Laboratory of Information Security, Institute of Information Engineering, Chinese Academy of Sciences, China (e-mail: rwq.renwenqi@gmail.com).

Runmin Cong is with the Institute of Information Science, Beijing Jiaotong University, Beijing 100044, China, and also with the Beijing Key Laboratory of Advanced Information Science and Network Technology, Beijing Jiaotong University, Beijing 100044, China (e-mail: rmcong@bjtu.edu.cn).

Junhui Hou and Sam Kwong are with the Department of Computer Science, City University of Hong Kong, Kowloon 999077, Hong Kong SAR, China, and also with the City University of Hong Kong Shenzhen Research Institute, Shenzhen 51800, China (e-mail: jh.hou@cityu.edu.hk; cssamk@cityu.edu.hk).

Dacheng Tao is with the UBTECH Sydney Artificial Intelligence Centre and the School of Information Technologies, the Faculty of Engineering
and Information Technologies, the University of Sydney, 6 Cleveland St, Darlington, NSW 2008, Australia (e-mail: dacheng.tao@sydney.edu.au).
}}
\markboth{IEEE TRANSACTIONS ON IMAGE PROCESSING}%
{Shell \MakeLowercase{\textit{et al.}}: Bare Demo of IEEEtran.cls for Journals}

\maketitle

\begin{abstract}
Underwater image enhancement has been attracting much attention due to its significance in marine engineering and aquatic robotics.
Numerous underwater image enhancement algorithms have been proposed in the last few years. However, these algorithms are mainly evaluated using either synthetic datasets or few selected real-world images. It is thus unclear how these algorithms would perform on images acquired in the wild and how we could gauge the progress in the field.
To bridge this gap, we present the first comprehensive perceptual study and analysis of underwater image enhancement using large-scale real-world images.
In this paper, we construct an \emph{Underwater Image Enhancement Benchmark (UIEB)} including 950 real-world underwater images, 890 of which have the corresponding reference images. We treat the rest 60 underwater images which cannot obtain satisfactory reference images as challenging data.
Using this dataset, we conduct a comprehensive study of the state-of-the-art underwater image enhancement algorithms qualitatively and quantitatively.
In addition, we propose an underwater image enhancement network (called Water-Net) trained on this benchmark as a baseline, which indicates the generalization of the proposed UIEB for training Convolutional Neural Networks (CNNs).
The benchmark evaluations and the proposed Water-Net demonstrate the performance and limitations of state-of-the-art algorithms, which shed light on future research in underwater image enhancement.
The dataset and code are available at \url{https://li-chongyi.github.io/proj_benchmark.html}.
\end{abstract}


\IEEEpeerreviewmaketitle

\section{Introduction}

\IEEEPARstart{D}{uring} the past few years, underwater image enhancement has drawn considerable attention in both image processing and underwater vision \cite{Jaffe2015,Sheinin2016}. Due to the complicated underwater environment and lighting conditions, enhancing underwater image is a challenging problem. Usually, an underwater image is degraded by wavelength-dependent absorption and scattering including forward scattering and backward scattering \cite{McGlamery1980,Jaffe1990,Hou2012,Akkaynak2017,Akkaynak20172}. In addition, the marine snow introduces noise and increases the effects of scattering. These adverse effects reduce visibility, decrease contrast, and even introduce color casts, which limit the practical applications of underwater images and videos in marine biology and archaeology \cite{Ludvigsen2007}, marine ecological \cite{Strachan1993}, to name a few
\cite{YangSurvey}. To solve this problem, earlier methods rely on multiple underwater images or polarization filters, while recent algorithms deal with this problem by using only information from a single image.

\begin{figure*}[!htb]
\centering
\includegraphics[width=18cm,height=4.5cm]{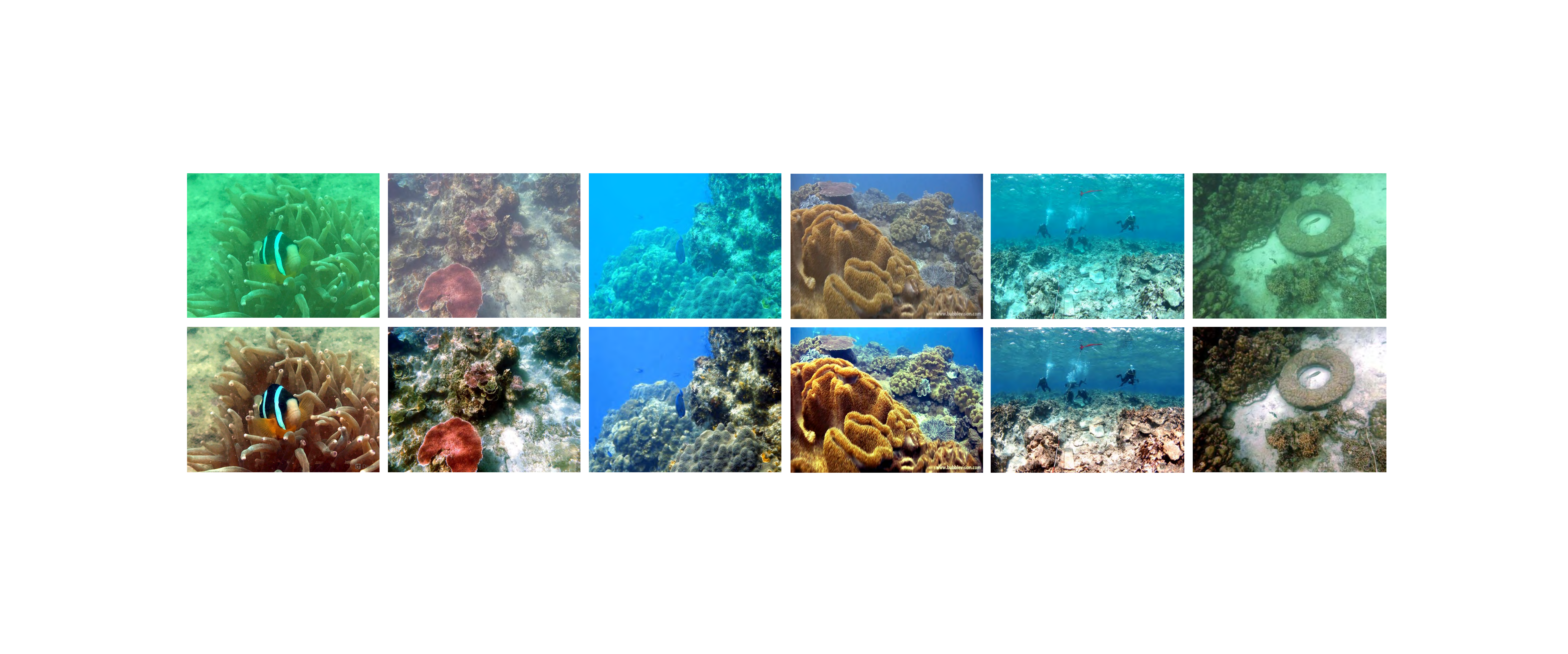}
\caption{Sampling images from UIEB. Top row: raw underwater images taken in diverse underwater scenes; Bottom row: the corresponding reference results.}
\label{fig_1}
\end{figure*}

Despite the prolific work, both the comprehensive study and insightful analysis of underwater image enhancement algorithms remain largely unsatisfactory due to the lack of a publicly available real-world underwater image dataset.
Additionally, it is practically impossible to simultaneously photograph a real underwater scene and the corresponding ground truth image for different water types.
Lacking sufficient and effective training data, the performance of deep learning-based underwater image enhancement algorithms does not match the success of recent deep learning-based high-level and low-level vision problems \cite{Dehazing,Survey,tgars,Ctc,Depth}.
To advance the development of underwater image enhancement, we construct a large-scale real-world \textbf{\emph{Underwater Image Enhancement Benchmark (UIEB)}}. Several sampling images and the corresponding reference images from UIEB are presented in Fig.~\ref{fig_1}. As shown, the raw underwater images in the UIEB have diverse color ranges and degrees of contrast decrease. In contrast, the corresponding reference images are color casts-free (at least relatively genuine color) and have improved visibility and brightness.
With the proposed UIEB, we carry out a comprehensive study for several state-of-the-art single underwater image enhancement algorithms both qualitatively and quantitatively, which enables insights into their performance and sheds light on future research.
In addition, with the constructed UIEB, CNNs can be easily trained to improve the visual quality of an underwater image. To demonstrate this application, we propose an underwater image enhancement model (Water-Net) trained by  the constructed UIEB.

The main contributions of this paper are summarized as follows.

\begin{itemize}
  \item We construct a large-scale real-world underwater image enhancement benchmark (\ie, \textbf{\emph{UIEB}}) which contains 950 real underwater images. These underwater images are likely taken under natural light, artificial light, or a mixture of natural light and artificial light. Moreover, the corresponding reference images for 890 images are provided according to laborious, time-consuming, and well-designed pairwise comparisons. UIEB provides a platform to evaluate, at least to some extent, the performance of different underwater image enhancement algorithms. It also makes supervised underwater image enhancement models which are out of the constraints of specific underwater scenes possible.

  \item  With the constructed UIEB, we conduct a comprehensive study of the state-of-the-art single underwater image enhancement algorithms ranging from qualitative to quantitative evaluations. Our evaluation and analysis provide comprehensive insights into the strengths and limitations of current underwater image enhancement algorithms, and suggest new research directions.

  \item We propose a CNN model (\ie, \emph{\textbf{Water-Net}}) trained by the UIEB for underwater image enhancement, which demonstrates the generalization of the constructed UIEB and the advantages of our Water-Net, and also motivates the development of deep learning-based underwater image enhancement.
\end{itemize}

\section{Existing Methodology, Evaluation Metric, and Dataset: An Overview}

\subsection{Underwater Image Enhancement Method}
Exploring underwater world has become an active issue in recent years \cite{Schettini2010,Han2018,Cui2017}. Underwater image enhancement as an indispensable step to improve the visual quality of recorded images has drawn much attention.  A variety of methods have been proposed and can be organized into four groups: supplementary information-based, non-physical model-based, physical model-based, and data-driven methods.

\noindent
\textbf{Supplementary Information-based Methods.} In the earlier stage, supplementary information from multiple images \cite{Narasimhan2003} or specialized hardware devices (\eg, polarization filtering \cite{Shashar1998,Schechner2004,Schechner2005,Treibitz2009}, range-gated imaging \cite{Tan2005,Li2009}, and fluorescence imaging \cite{Treibitz2012,Murez2015}) were utilized to improve the visibility of underwater images. Compared to supplementary information-based methods, single underwater image enhancement has been proven to be more suitable for challenging situations such as dynamic scenes, and thus, gains extensive attention.

\noindent
\textbf{Non-physical Model-based Methods.} Non-physical model-based methods aim to modify image pixel values to improve visual quality. Iqbal \etal~\cite{Iqbal2010} stretched the dynamic pixel range in RGB color space and HSV color space to improve the contrast and saturation of an underwater image.
Chani and Isa \cite{Ghani2015,Ghani20152} modified the work of \cite{Iqbal2010} to reduce the over-/under-enhanced regions by shaping the stretching process following the Rayleigh distribution.
Ancuti \etal \cite{Ancuti2012} proposed an underwater image enhancement method by blending a contrast-enhanced image and a color-corrected image in a multi-scale fusion strategy.
In~\cite{Fu2017}, a two-step approach for underwater image enhancement was proposed, which includes a color correction algorithm and a contrast enhancement algorithm.

Another line of research tries to enhance underwater images based on the Retinex model.
Fu \etal~\cite{Fu2014} proposed a retinex-based method for underwater image enhancement, which consists of color correction, layer decomposition, and enhancement.
Zhang \etal~\cite{Zhang2017Neuro} proposed an extended multi-scale retinex-based underwater image enhancement method. In this work, the underwater

\noindent
\textbf{Physical Model-based Methods.} Physical model-based methods regard the enhancement of an underwater image as an inverse problem, where the latent parameters of an image formation model are estimated from a given image.
These methods usually follow the same pipeline: 1) building a physical model of the degradation; 2) estimating the unknown model parameters; and 3) addressing this inverse problem.

One line of research is to modify the Dark Channel Prior (DCP) \cite{He2011} for underwater image enhancement.
In \cite{Chiang2012}, DCP was combined with the wavelength-dependent compensation algorithm to restore underwater images.
In \cite{Drews2016}, an Underwater Dark Channel Prior (UDCP) was proposed based on the fact that the information of the red channel in an underwater image is undependable.
Based on the observation that the dark channel of the underwater image tends to be a zero map, Liu and Chau \cite{Liu2016} formulated a cost function and minimized it so as to find the optimal transmission map, which is able to maximize the image contrast.
Instead of the DCP, Li \etal~\cite{Li2017prl} employed the random forest regression model to estimate the transmission of the underwater scenes.
Recently, Peng \etal~\cite{Peng2018} proposed a Generalized Dark Channel Prior (GDCP) for image restoration, which incorporates adaptive color correction into an image formation model.

Another line of research try to employ the optical properties of underwater imaging.
Carlevaris-Bianca \etal~\cite{Carlevaris2010} proposed a prior that exploits the difference in attenuation among three color channels in RGB color space to predict the transmission of an underwater scene. The idea behind this prior is that the red light usually attenuates faster than the green light and the blue light in an underwater scenario.
Galdran \etal \cite{Galdran2015} proposed a Red Channel method, which recovers the lost contrast of an underwater image by restoring the colors associated with short wavelengths.
According to the findings that the background color of underwater images has relations with the inherent optical properties of water medium, Zhao \etal~\cite{Zhao2015} enhanced the degraded underwater images by deriving inherent optical properties of water from the background color.
Li \etal~\cite{Li2016ICIP,Li2016} proposed an underwater image enhancement method based on the minimum information loss principle and histogram distribution prior.
Peng \etal~\cite{Peng2017} proposed a depth estimation method for underwater scenes based on image blurriness and light absorption, which is employed to enhance underwater images.
Berman \etal~\cite{Dana2017} took multiple spectral profiles of different water types into account and reduced the problem of underwater image restoration to single image dehazing.
Wang \etal~\cite{Wang2018} combined the adaptive attenuation-curve prior with the characteristics of underwater light propagation for underwater image restoration.
More recently, Akkaynak and Treibitz \cite{Akkaynak2019} proposed an underwater image color correction method based on a revised underwater image formation model \cite{Akkaynak2017} which is physically accurate. The existing physical model-based methods, except for the recent work \cite{Akkaynak2019}, follow the simplified image formation models that assume the attenuation coefficients are only properties of the water and are uniform across the scene per color channel. This assumption leads to the unstable and visually unpleasing results as demonstrated in \cite{Akkaynak2017,Akkaynak2019}.

\noindent
\textbf{Data-driven Methods.} Recent years have witnessed the significant advance of deep learning in low-level vision problems. These methods can be trained using synthetic pairs of degraded images and high-quality counterparts.  However, underwater image formation models depend on specific scenes and lighting conditions, and even are related to temperature and turbidity. Thus, it is difficult to synthesize realistic underwater images for CNNs training. Further, the learned distribution by CNNs trained on synthetic underwater images does not always generalize to real-world cases. Therefore, the performance and the amount of deep learning-based underwater image enhancement methods do not match the success of recent deep learning-based low-level vision problems \cite{Survey2019}.

Recently, Li \etal~\cite{WaterGAN} proposed a deep learning-based underwater image enhancement model, called WaterGAN. WaterGAN first simulates underwater images from the in-air image and depth pairings in an unsupervised pipeline. With the synthetic training data, the authors use a two-stage network for underwater image restoration, especially for color casts removal.
Underwater image enhancement models (\ie, UWCNNs) trained by ten types of underwater images was proposed in~\cite{UWCNN}, where underwater images are synthesized based on a revised underwater image formation model \cite{Akkaynak2017} and the corresponding underwater scene parameters.
More recently, a weakly supervised underwater color transfer model \cite{Emerging} (\ie, Water CycleGAN) was proposed based on Cycle-Consistent Adversarial Networks \cite{Cycle}. Benefiting from the adversarial network architecture and multi-term loss function, this network model relaxes the need for paired underwater images for training and allows the underwater images being taken in unknown locations. However, it tends to produce inauthentic results in some cases due to the nature of multiple possible outputs.
Guo \etal~\cite{Guo2019} proposed a multiscale dense GAN (\ie, Dense GAN) for underwater image enhancement. The authors combined the non-saturating GAN loss with the $\ell_{1}$ loss and gradient loss to learn the distribution of ground truth images in the feature domain. However, this method still cannot avoid the limitations of multiple possible outputs from GANs.
Therefore, the robustness and generalization of deep learning-based underwater enhancement methods still fall behind conventional state-of-the-art methods.

\subsection{Underwater Image Quality Evaluation}

In the following, we will give a brief introduction of the image quality evaluation metrics which are widely used for underwater image enhancement methods.

\noindent
\textbf{Full-reference Metrics.} For an underwater image with ground truth image, the full-reference image quality evaluation metrics (\eg, MSE, PSNR, and SSIM \cite{SSIM}) were employed for evaluations. Such underwater images usually are a few color checker images or color image patches taken in the simulated or real underwater environment. For example, Zhao \etal~\cite{Zhao2015} treated a plastic color disk as ground truth image and captured its underwater image in a water pool as the testing image.

\noindent
\textbf{Non-reference Metrics.} Different from other low-level vision problems where the ground truth images can be easily obtained (\eg, image super-resolution), it is challenging to achieve a large amount of paired underwater images and the corresponding ground truth images. For a real-world underwater image where the ground truth image was unavailable, non-reference image quality evaluation metrics, such as image entropy, visible edges \cite{Hautiere2011}, and dynamic range independent image quality assessment \cite{Aydin2008}, were utilized.
In addition, some authors employed specific applications, like feature point matching, edge detection, and image segmentation, to evaluate their results. Besides, several specific non-reference metrics were proposed for underwater image quality evaluation. Yang and Sowmya~\cite{Yang2015} proposed an underwater color image quality evaluation metric (\ie, UCIQE). UCIQE first quantifies the non-uniform color casts, blurring, and low contrast, and then combines these three components in a linear manner.  In \cite{Panetta2016}, the authors proposed a non-reference underwater image quality measure, called UIQM, which comprises three attribute measures: colorfulness measure, sharpness measure, and contrast measure. Each presented attribute measure is inspired by the properties of human visual system.

\subsection{Underwater Image Datasets}

There are several real-world underwater image datasets such as Fish4Knowlege dataset for underwater target detection and recognition\footnote[1]{\url{http://groups.inf.ed.ac.uk/f4k/}}, underwater images in SUN dataset for scene recognition and object detection\footnote[2]{\url{http://groups.csail.mit.edu/vision/SUN/}}\cite{SUN}, MARIS dataset for marine autonomous robotics\footnote[3]{\url{http://rimlab.ce.unipr.it/Maris.html}}, Sea-thru dataset including 1100 underwater image with range maps\footnote[4]{\url{http://csms.haifa.ac.il/profiles/tTreibitz/datasets/sea_thru/index.html}}\cite{Akkaynak2019}, and Haze-line dataset providing raw images, TIF files, camera calibration files, and distance maps\footnote[5]{\url{http://csms.haifa.ac.il/profiles/tTreibitz/datasets/ambient_forwardlooking/index.html}}\cite{BermanData}.
However, existing datasets usually have monotonous content and limited scenes, few degradation characteristics, and insufficient data. Moreover, these datasets did not provide the corresponding ground truth images or reference results since it is difficult or even impractical to simultaneously obtain a real underwater image and the corresponding ground truth image of the same scene due to the diverse water types and lighting conditions as well as expensive and logistically complex imaging devices. In recent years, several underwater image synthesis methods have been proposed. Li \etal~\cite{WaterGAN} proposed a GAN-based method\footnote[6]{\url{https://github.com/kskin/WaterGAN}} while Duarte \etal~\cite{Duarte2016} simulated underwater image degradation using milk, chlorophyll, or green tea in a tank.
Blasinski \etal~\cite{Blasinski2017} provided an open-source underwater image simulation tool and a three parameter underwater image formation model \cite{Blasinski2016}. Li \etal~\cite{UWCNN} proposed a synthetic underwater image dataset including ten subsets for different types of water\footnote[7]{\url{https://li-chongyi.github.io/proj_underwater_image_synthesis.html}}. However, there still exists a gap between synthetic and real-world underwater images. Therefore, it is challenging to evaluate the state-of-the-art methods fairly and comprehensively, and is hard to develop effective deep learning-based models.

\section{Proposed Benchmark Dataset}

After systematically reviewing previous work, we found the main issue existing in the community of underwater image enhancement is lacking a large-scale real-world underwater image dataset with reference images. In what follows, we introduce the constructed dataset in detail, including data collection and reference image generation.

\subsection{Data Collection}

\begin{figure}[!htb]
  \centering
\begin{minipage}[b]{0.3\linewidth}
  \centering
  \centerline{\includegraphics[width=7cm,height=4cm]{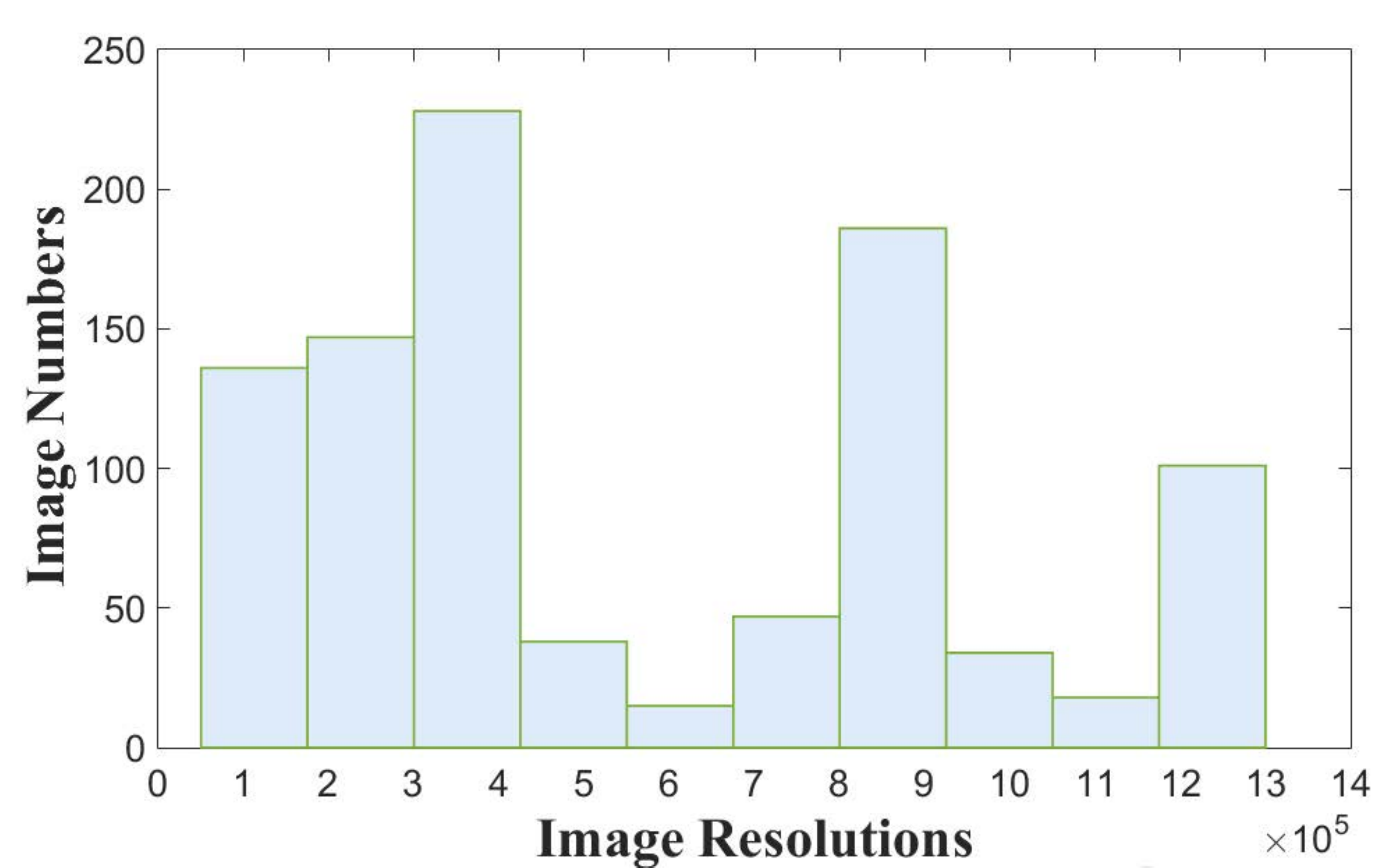}}
  \centerline{(a)}\medskip
\end{minipage}

 \begin{minipage}[b]{0.3\linewidth}
  \centering
  \centerline{\includegraphics[width=7cm,height=4cm]{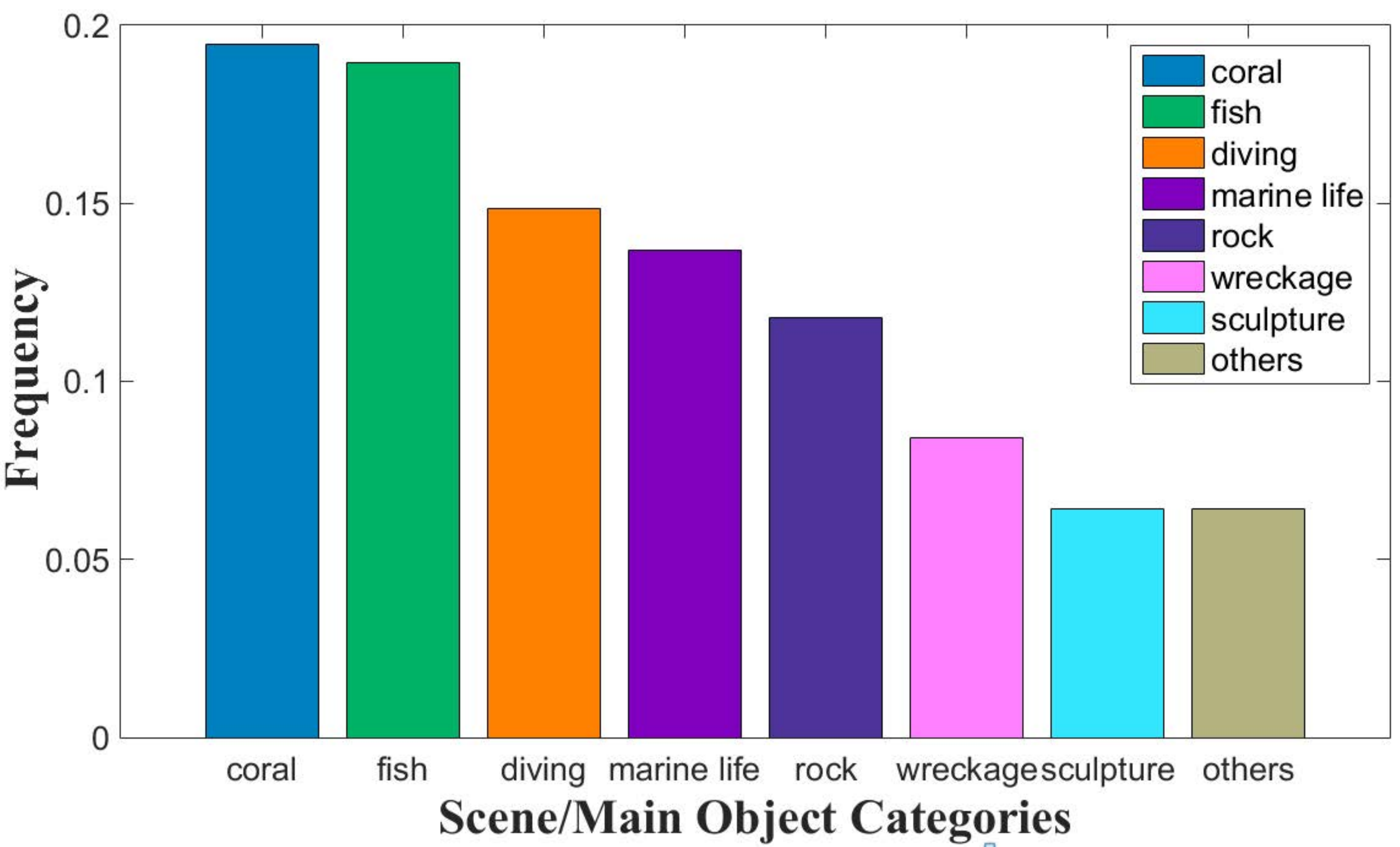}}
  \centerline{(b)}\medskip
\end{minipage}
 \caption{Statistics of the constructed UIEB. (a) Image resolutions. (b) Scene/main object categories.}
\label{fig:Statistics}
\end{figure}

There are three objectives for underwater image collection:

\begin{enumerate}
  \item a diversity of underwater scenes, different characteristics of quality degradation, and a broad range of image content should be covered;
  \item the amount of underwater images should be large; and
  \item the corresponding high-quality reference images should be provided so that pairs of images enable fair image quality evaluation and end-to-end learning.
\end{enumerate}

\begin{figure}[!htb]
\centering
\includegraphics[width=8cm,height=7cm]{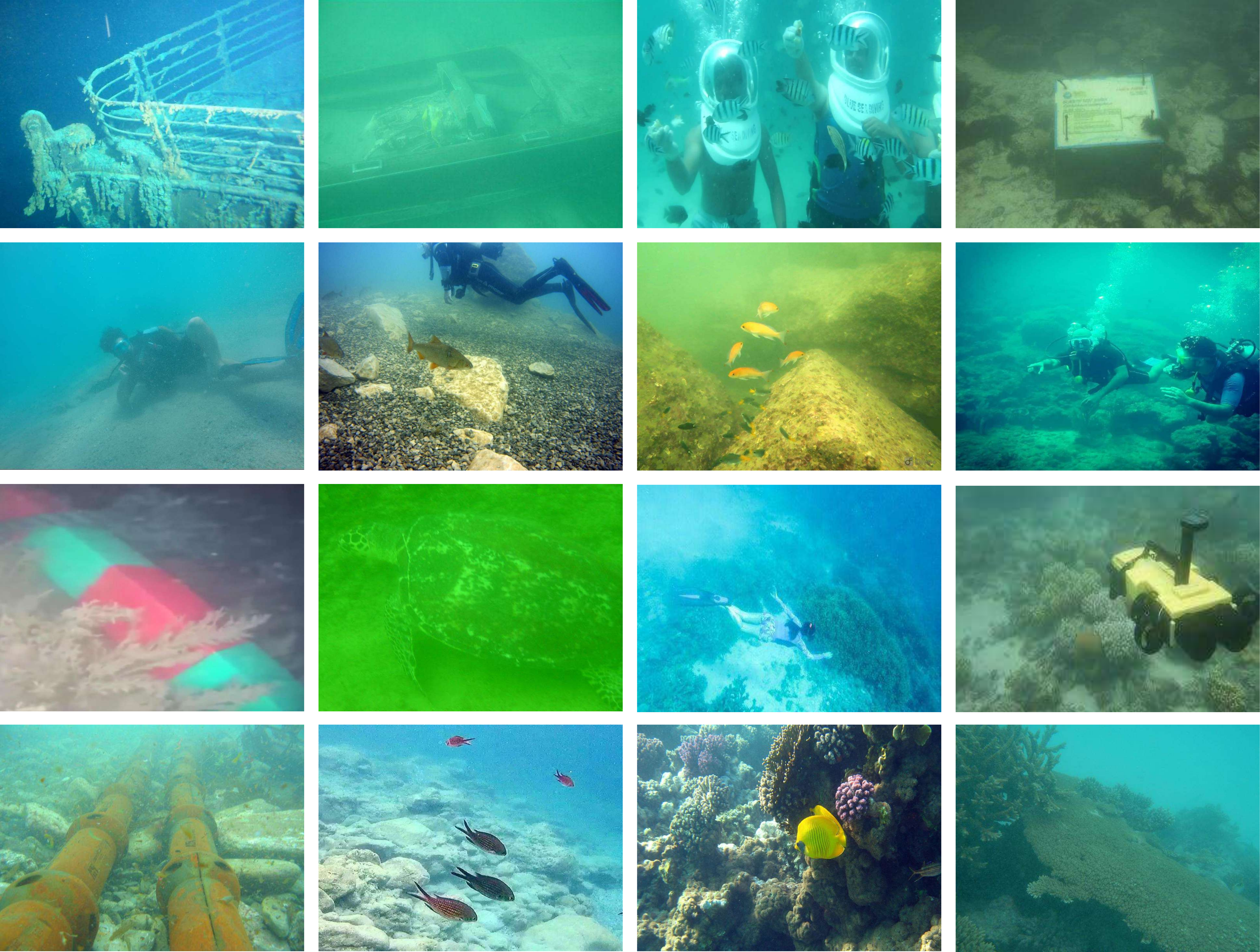}
\caption{Examples of the images in UIEB. These images have obvious characteristics of underwater image quality degradation (\eg, color casts, decreased contrast, and blurring details) and are taken in a diversity of underwater scenes.}
\label{fig_2}
\end{figure}

To achieve the first two objectives, we first collect a large number of underwater images, and then refine them. These underwater images
are collected from Google, YouTube, related papers \cite{Treibitz2009,Ancuti2012,Fu2017,Fu2014,Galdran2015}, and our self-captured videos. We mainly retain the underwater images which meet the first objective. After data refinement, most of the collected images are weeded out, and about 950 candidate images are remaining. We provide a statistic of image resolutions and the scene/main object categories of the UIEB in Fig.~\ref{fig:Statistics} and present some examples of the images in Fig.~\ref{fig_2}.

As shown in Figs.~\ref{fig:Statistics} and~\ref{fig_2}, the UIEB contains a large range of image resolutions and spans diverse scene/main object categories including coral (\eg, fringing reefs and barrier reefs), marine life (\eg, turtles and sharks), \etc. To achieve the third objective, we introduce a high-quality reference image generation method in the next section.

\subsection{Reference Image Generation}

With the candidate underwater images, the potential reference images are generated by 12 image enhancement methods, including 9 underwater image enhancement methods (\ie, fusion-based \cite{Ancuti2012}, two-step-based \cite{Fu2017}, retinex-based \cite{Fu2014}, UDCP \cite{Drews2016}, regression-based~\cite{Li2017prl}, GDCP \cite{Peng2018}, Red Channel \cite{Galdran2015}, histogram prior~\cite{Li2016}, and blurriness-based~\cite{Peng2017}), 2 image dehazing methods (\ie, DCP~\cite{He2011} and MSCNN~\cite{Ren2016}), and 1 commercial application for enhancing underwater images (\ie, dive+\footnote[8]{\url{https://itunes.apple.com/us/app/dive-video-color-correction/id1251506403?mt=8}}). We exclude the recent deep learning-based methods due to their limited generalization capability to the diverse real-world underwater images and the fixed size of network output \cite{Emerging,Guo2019}. The source codes of all the employed methods except the fusion-based \cite{Ancuti2012} are provided by their authors. We reimplement the fusion-based \cite{Ancuti2012} since the source code is unavailable. For the dive+, we tune its parameter settings to generate satisfactory results. At last, we totally generate 12 $\times$ 950 enhanced results.

\begin{figure}[!htb]
\centering
\includegraphics[width=8.5cm,height=2.5cm]{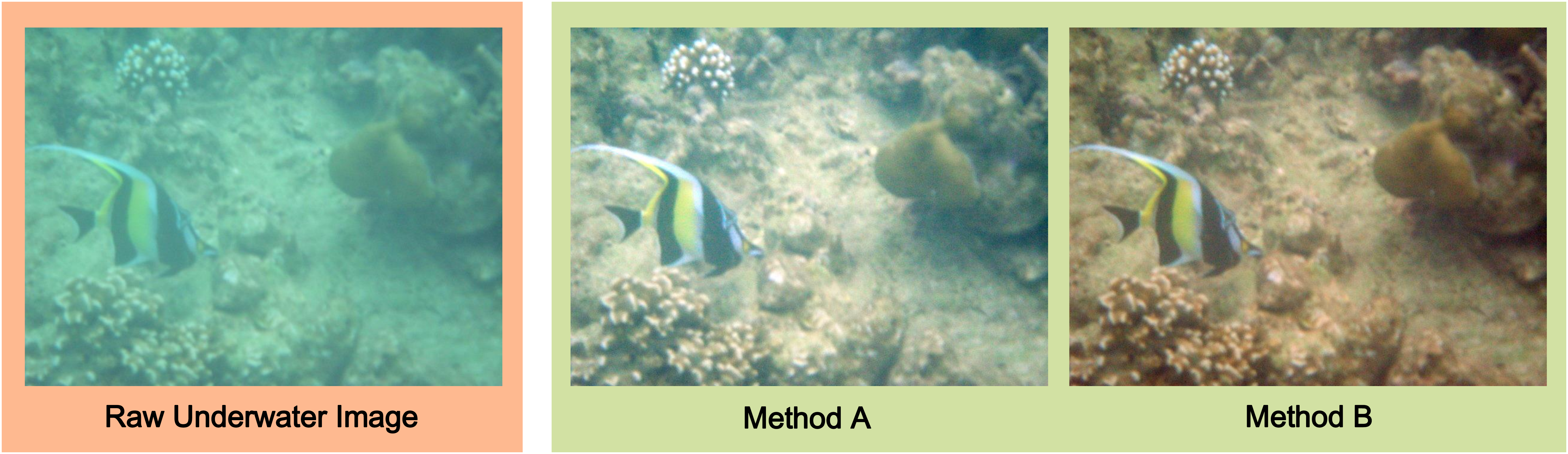}
\caption{An example of pairwise comparisons. Treating the raw underwater image as a reference, a volunteer needs to independently decide which one is better between the results of method A and method B.}
\label{fig_3}
\end{figure}

\begin{figure*}[!htb]
\centering
\includegraphics[width=18cm,height=5cm]{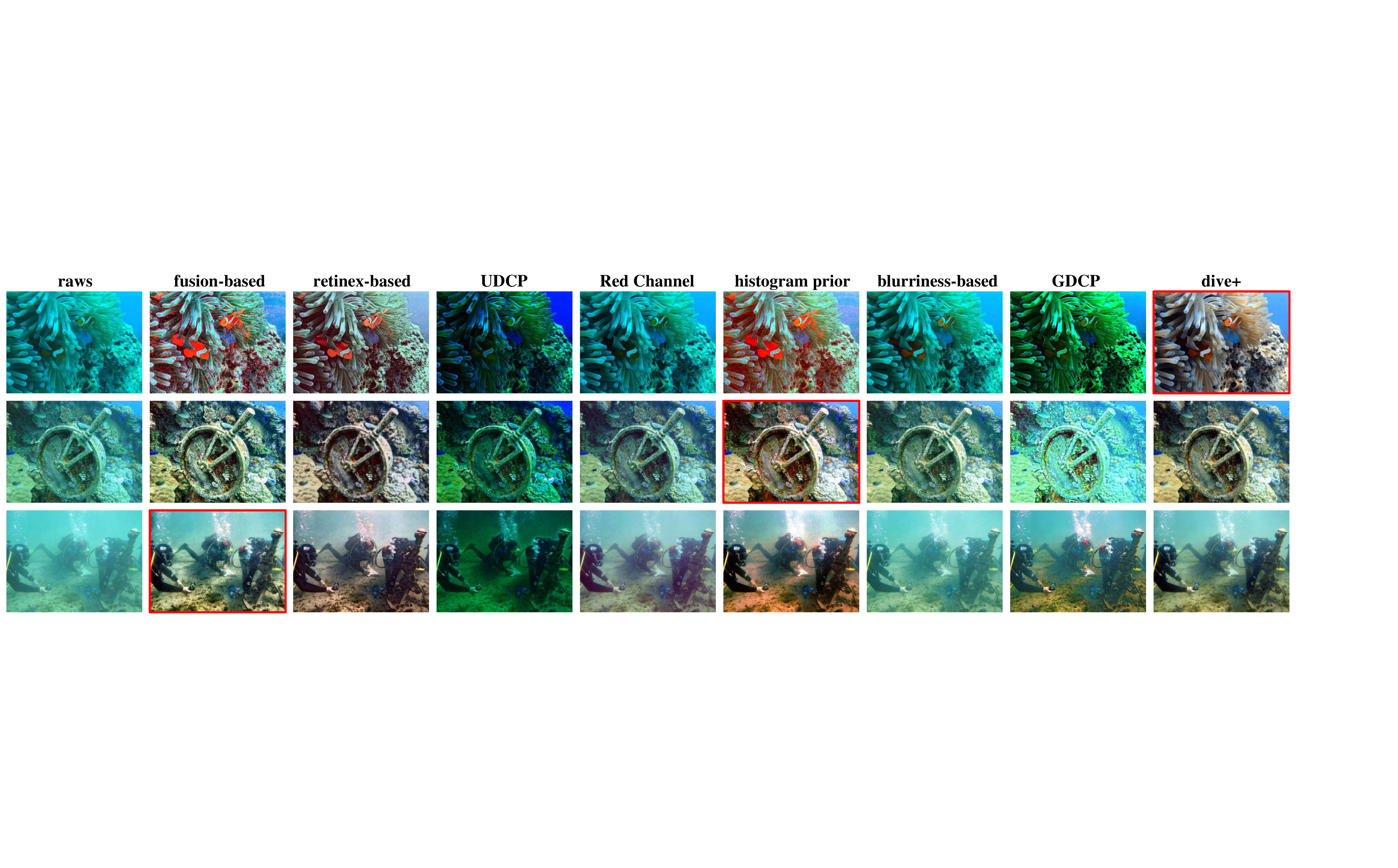}
\caption{Results generated by different methods. From left to right are raw underwater images, and the results of fusion-based \cite{Ancuti2012}, retinex-based \cite{Fu2014}, UDCP \cite{Drews2016}, Red Channel \cite{Galdran2015}, histogram prior \cite{Li2016}, blurriness-based~\cite{Peng2017}, GDCP~\cite{Peng2018} and dive+. Red boxes indicate the final reference images.}
\label{fig_4}
\end{figure*}

With raw underwater images and the enhanced results, we invite 50 volunteers (25 volunteers with image processing experience; 25 volunteers without related experience) to perform pairwise comparisons among the 12 enhanced results of each raw underwater image under the same monitor.

Specifically, each volunteer is shown a raw underwater image and a set of enhanced result pairs. The enhanced image pairs are drawn from all the competitive methods randomly, and the result winning the pairwise comparisons will be compared again in the next round, until the best one is selected. There is no time constraint for volunteers and zoom-in operation is allowed. An example of pairwise comparisons is shown in Fig.~\ref{fig_3}. For each pair of enhanced results, taking the raw underwater image as a reference, a volunteer first needs to independently decide which one is better than the other. For each volunteer, the best result will be selected after 11 pairwise comparisons.

\begin{table}[htbp]
\caption{Percentage of the reference images from the results of different methods. }
 \centering
\begin{tabular}{c|c}
  \hline
  \textbf{Method} & \textbf{Percentage (\%)} \\
  \hline
  fusion-based \cite{Ancuti2012} & \textcolor[rgb]{0.00,0.07,1.00}{24.72} \\
  two-step-based \cite{Fu2017} & 7.30  \\
  retinex-based \cite{Fu2014} & 0.22 \\
  DCP~\cite{He2011}  & 2.58 \\
  UDCP \cite{Drews2016} & 0.00 \\
  regression-based~\cite{Li2017prl} & 1.80\\
  GDCP~\cite{Peng2018} & 0.34 \\
  Red Channel \cite{Galdran2015} & 0.90 \\
  histogram prior~\cite{Li2016} & 13.37 \\
  blurriness-based~\cite{Peng2017} & 3.48 \\
  MSCNN~\cite{Ren2016} & 0.90\\
  dive+ &  \textcolor[rgb]{1.00,0.00,0.00}{43.93}\\
  \hline
\end{tabular}
\vspace{\baselineskip}
\label{table_1}
\end{table}

Additionally, the volunteer needs to inspect the best result again and then label the best result as being satisfactory or dissatisfactory. The reference image for a raw underwater image is first selected by majority voting after pairwise comparisons. After that, if the selected reference image has greater than half the number of votes labeled dissatisfaction, its corresponding raw underwater image is treated as a challenging image and the reference image is discarded. We totally achieve 890 available reference images which have higher quality than any individual methods and a challenging set including 60 underwater images. To visualize the process of reference image generation, we present some cases that the results of some methods are shown and indicate which one is the final reference image in Fig.~\ref{fig_4}. Furthermore, the percentage of the reference images from the results of different methods is presented in Table~\ref{table_1}. In this paper, we highlight the top one performance in red, whereas the second top one is in blue.

In summary, the results with improved contrast and genuine color are most favored by observers while the over-/under- enhancement, artifacts, and color casts lead to visually unpleasing results. Finally, the constructed UIEB includes two subsets: 890 raw underwater images with the corresponding high-quality reference images; 60 challenging underwater images. To the best of our knowledge, it is the first real-world underwater image dataset with reference images so far. The UIEB has various potential applications, such as performance evaluation and CNNs training. Next, we will introduce these two applications.

\section{Evaluation and Discussion}

A comprehensive and fair evaluation of underwater image enhancement methods has long been missing from the literatures. Using the constructed UIEB, we evaluate the state-of-the-art underwater image enhancement methods (\ie, fusion-based \cite{Ancuti2012}, two-step-based \cite{Fu2017}, retinex-based \cite{Fu2014}, UDCP \cite{Drews2016}, regression-based~\cite{Li2017prl}, GDCP \cite{Peng2018}, Red Channel \cite{Galdran2015}, histogram prior~\cite{Li2016}, blurriness-based~\cite{Peng2017}) both qualitatively and quantitatively.

\subsection{Qualitative Evaluation}

\begin{figure*}[!htb]
\centering
\includegraphics[width=18cm,height=3cm]{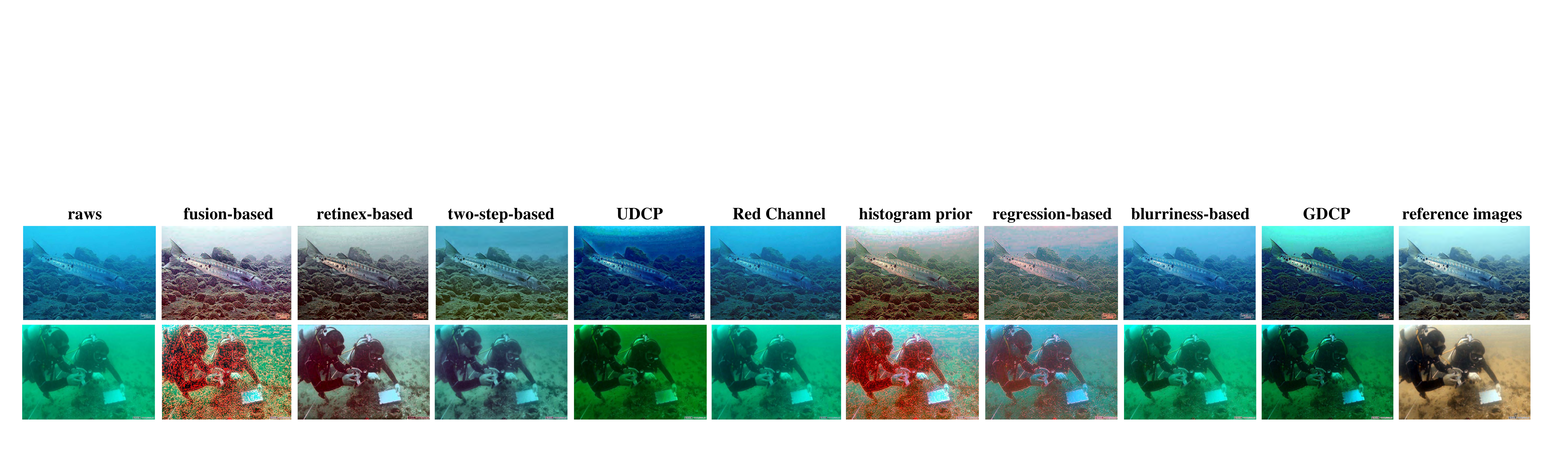}
\caption{Subjective comparisons on bluish and greenish  underwater images. From left to right are raw underwater images, and the results of fusion-based \cite{Ancuti2012}, retinex-based \cite{Fu2014}, two-step-based \cite{Fu2017}, UDCP \cite{Drews2016}, Red Channel \cite{Galdran2015}, histogram prior~\cite{Li2016}, regression-based~\cite{Li2017prl}, blurriness-based~\cite{Peng2017}, GDCP~\cite{Peng2018}, and reference images.}
\label{fig_5}
\end{figure*}

\begin{figure*}[!htb]
\centering
\includegraphics[width=18cm,height=3cm]{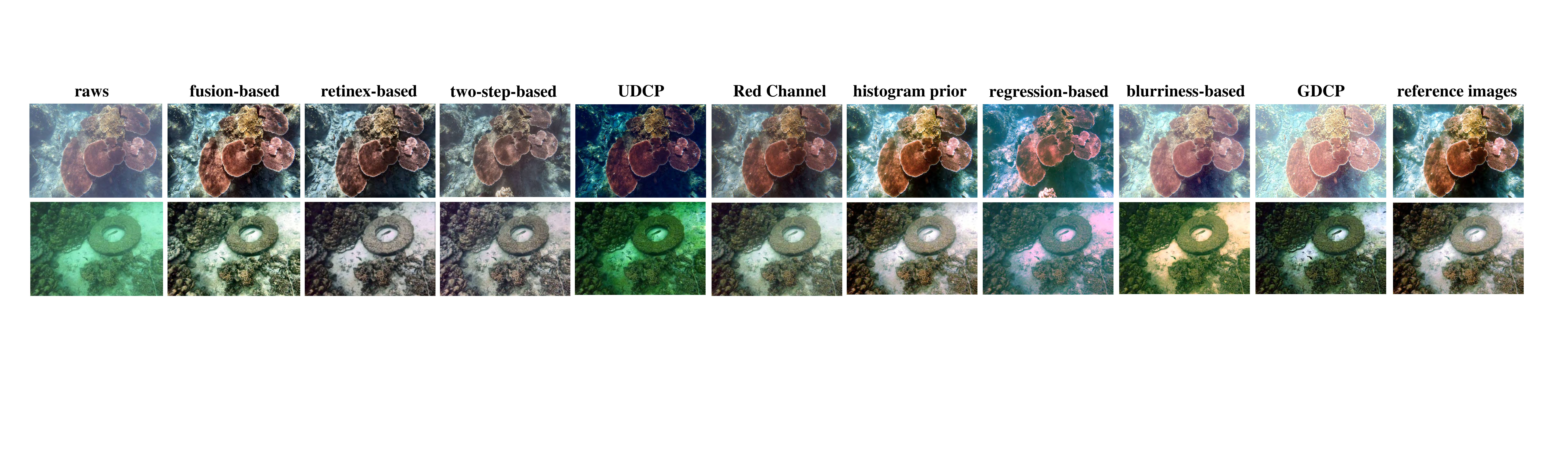}
\caption{Subjective comparisons on downward looking images. From left to right are raw underwater images, and the results of fusion-based \cite{Ancuti2012}, retinex-based \cite{Fu2014}, two-step-based \cite{Fu2017}, UDCP \cite{Drews2016}, Red Channel \cite{Galdran2015}, histogram prior~\cite{Li2016}, regression-based~\cite{Li2017prl}, blurriness-based~\cite{Peng2017}, GDCP~\cite{Peng2018}, and reference images.}
\label{fig_6}
\end{figure*}

\begin{figure*}[!htb]
\centering
\includegraphics[width=18cm,height=3cm]{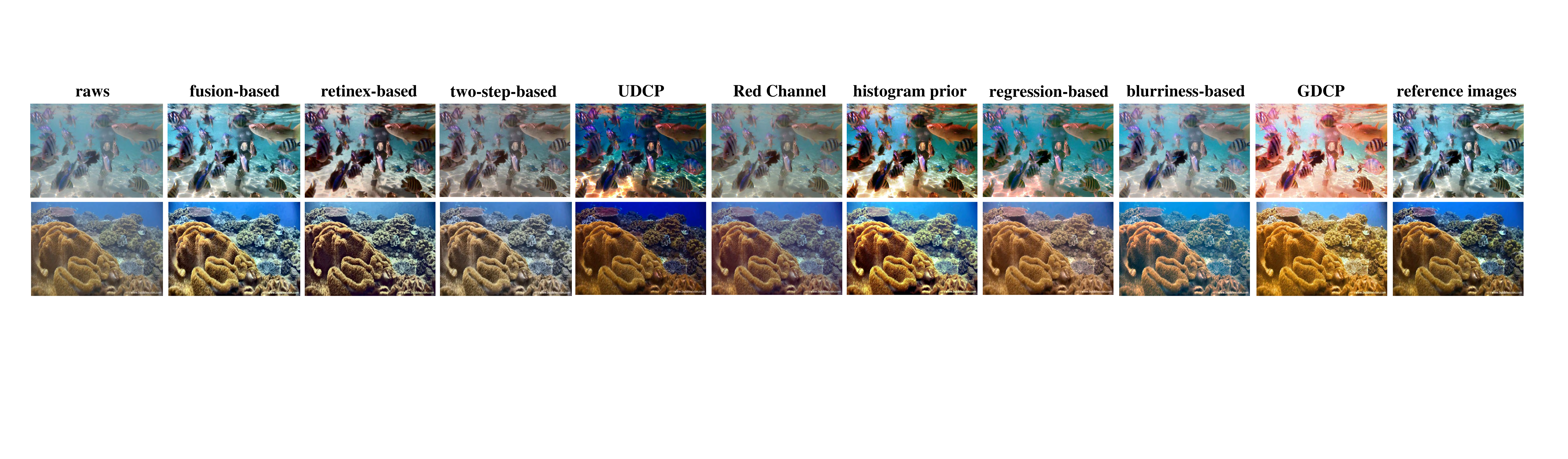}
\caption{Subjective comparisons on forward looking images. From left to right are raw underwater images, and the results of fusion-based \cite{Ancuti2012}, retinex-based \cite{Fu2014}, two-step-based \cite{Fu2017}, UDCP \cite{Drews2016}, Red Channel \cite{Galdran2015}, histogram prior~\cite{Li2016}, regression-based~\cite{Li2017prl}, blurriness-based~\cite{Peng2017}, GDCP~\cite{Peng2018}, and reference images.}
\label{fig_7}
\end{figure*}

\begin{figure*}[!htb]
\centering
\includegraphics[width=18cm,height=3cm]{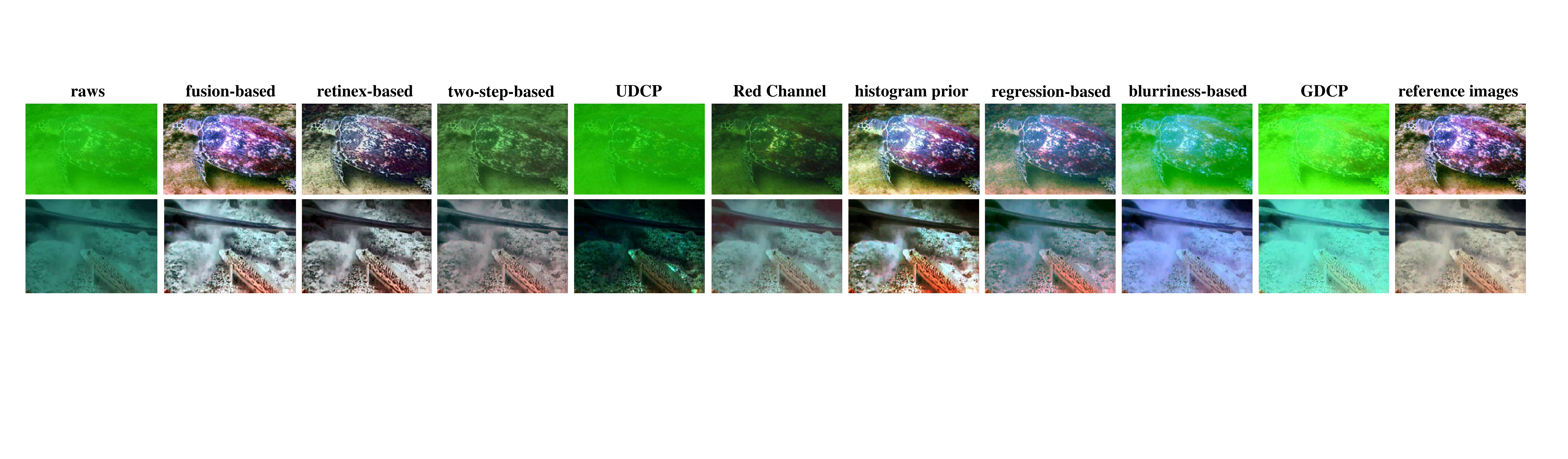}
\caption{Subjective comparisons on low backscatter scenes. From left to right are raw underwater images, and the results of fusion-based \cite{Ancuti2012}, retinex-based \cite{Fu2014}, two-step-based \cite{Fu2017}, UDCP \cite{Drews2016}, Red Channel \cite{Galdran2015}, histogram prior~\cite{Li2016}, regression-based~\cite{Li2017prl}, blurriness-based~\cite{Peng2017}, GDCP~\cite{Peng2018}, and reference images.}
\label{fig_8}
\end{figure*}

\begin{figure*}[!htb]
\centering
\includegraphics[width=18cm,height=3cm]{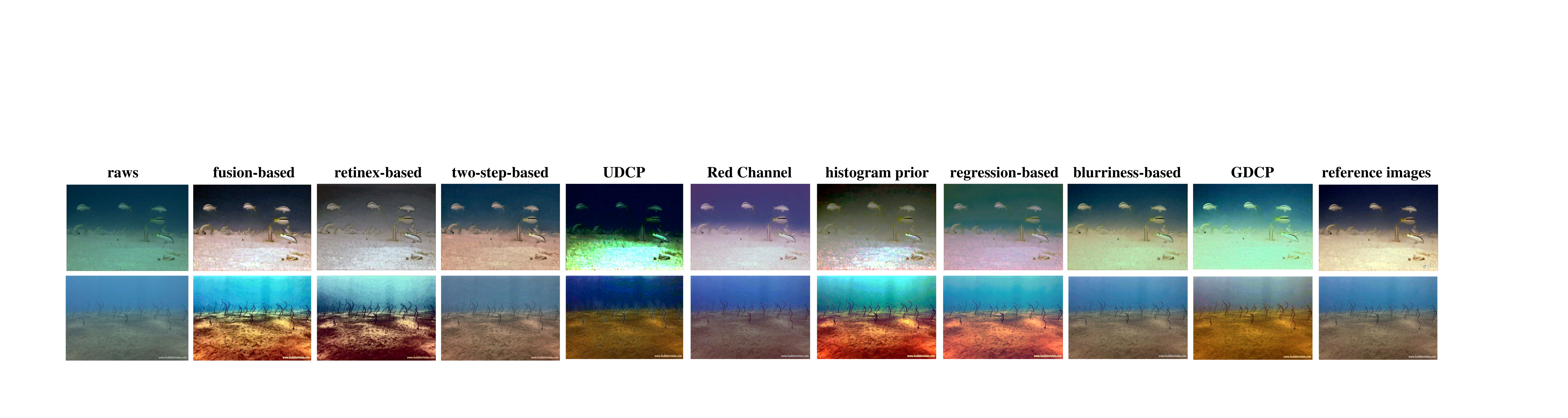}
\caption{Subjective comparisons on high backscatter scenes. From left to right are raw underwater images, and the results of fusion-based \cite{Ancuti2012}, retinex-based \cite{Fu2014}, two-step-based \cite{Fu2017}, UDCP \cite{Drews2016}, Red Channel \cite{Galdran2015}, histogram prior~\cite{Li2016}, regression-based~\cite{Li2017prl}, blurriness-based~\cite{Peng2017}, GDCP~\cite{Peng2018}, and reference images.}
\label{fig_99}
\end{figure*}

We first select several underwater images from the UIEB, and then divide these images into five categories: greenish and bluish images, downward looking images, forward looking images, low backscatter scenes, and high backscatter scenes. The results of different methods and the corresponding reference images are shown in Figs.~\ref{fig_5}-\ref{fig_99}. Best viewed with zoom-in on a digital display. Note that these underwater images cannot cover the entire UIEB.

In open water, the red light first disappears because of its longest wavelength, followed by the green light and then the blue light \cite{Akkaynak20172}. Such selective attenuation in open water results in bluish or greenish underwater images, such as the raw underwater images in Fig.~\ref{fig_5}. Color deviation seriously affects the visual quality of underwater images and is difficult to be removed. As shown, the fusion-based \cite{Ancuti2012}, histogram prior~\cite{Li2016}, and regression-based~\cite{Li2017prl} introduce reddish color deviation due to the inaccurate color correction algorithms used in \cite{Ancuti2012,Li2017prl} and the histogram distribution prior \cite{Li2016}. The retinex-based \cite{Fu2014} removes the color deviation well, while the UDCP \cite{Drews2016} and GDCP~\cite{Peng2018} aggravate the effect of color casts. The two-step-based \cite{Fu2017} can effectively increase the contrast of underwater images. The Red Channel \cite{Galdran2015} and blurriness-based~\cite{Peng2017} have less positive effect on these two images on account of the limitations of the priors used in these two methods.

For the downward looking images and forward looking images shown in Figs.~\ref{fig_6} and~\ref{fig_7}, the fusion-based \cite{Ancuti2012}, retinex-based \cite{Fu2014}, and histogram prior~\cite{Li2016} significantly remove the effect of haze on the underwater images while the  two-step-based \cite{Fu2017}, Red Channel \cite{Galdran2015}, and blurriness-based~\cite{Peng2017} remain some haze in the results. We note that UDCP \cite{Drews2016}, regression-based~\cite{Li2017prl}, and  GDCP~\cite{Peng2018} tend to bring in color deviation in the enhanced results. In terms of the physical-model based methods (\eg,  UDCP \cite{Drews2016}, Red Channel \cite{Galdran2015}, regression-based~\cite{Li2017prl}, blurriness-based~\cite{Peng2017}, and GDCP~\cite{Peng2018}), it is hard to estimate the veiling light from the downward looking images accurately. In addition, the physical-model based methods may incorrectly estimate the veiling light from the RGB values of water in textures regions of forward looking images

Backscatter reduces the contrast and produces the foggy veiling in an underwater image. The effect of backscatter is related to the distance between the camera and the scene. For the low backscatter scenes (short distance between the camera and the scene) in Fig.~\ref{fig_8}, the effect of backscatter is relatively easy to be removed. In contrast, the high backscatter (long distance) significantly degrades the visual quality of underwater images. As shown in Fig.~\ref{fig_99}, all the physical model-based methods (\eg,  UDCP \cite{Drews2016}, Red Channel \cite{Galdran2015}, regression-based~\cite{Li2017prl}, blurriness-based~\cite{Peng2017}, and GDCP~\cite{Peng2018}) cannot remove the high backscatter due to the inaccurate physical models and assumptions used in these methods.

In summary, the fusion-based \cite{Ancuti2012} has relatively decent performance on a variety of underwater images. The method of UDCP \cite{Drews2016} tends to produce artifacts on enhanced results in some cases. Other competitors are effective to some extent. In fact, it is almost impossible for a color correction algorithm or a kind of prior effective for all types of underwater images. Moreover, the effect of high backscatter is challenging for underwater image enhancement.

\subsection{Quantitative Evaluation}

To quantitatively evaluate the performance of different methods, we perform the full-reference evaluation, non-reference evaluation, and runtime evaluation.

\subsubsection{Full-reference Evaluation}

We first conduct a full-reference evaluation using three commonly-used metrics (\ie, MSE, PSNR, and SSIM). The results of full-reference image quality evaluation by using the reference images can provide realistic feedback of the performance of different methods to some extent, although the real ground truth images might be different from the reference images.
A higher PSNR score and a lower MSE score denote the result is closer to the reference image in terms of image content, while a higher SSIM score means the result is more similar to the reference image in terms of image structure and texture.
We present the average scores of different methods on the 890 images with reference images in the UIEB. As shown in Table~\ref{table_2}, the dive+ stands out as the best performer across all metrics. In addition, the fusion-based \cite{Ancuti2012} ranks the second best in terms of the full-reference metrics. It is reasonable for such results, since most reference images are selected from the results generated by dive+ and fusion-based \cite{Ancuti2012}.

\begin{table}[htbp]
\caption{Full-reference Image Quality Evaluation.}
 \centering
\begin{tabular}{c|c|c|c}
  \hline
  \textbf{Method} & \textbf{MSE ($\times 10^3$)} $\downarrow$ & \textbf{PSNR (dB)} $\uparrow$ & \textbf{SSIM} $\uparrow$ \\
  \hline
  fusion-based \cite{Ancuti2012} & {\color{blue}0.8679} & {\color{blue}18.7461} & {\color{blue}0.8162}\\
  two-step-based \cite{Fu2017} & 1.1146 &17.6596 &0.7199 \\
  retinex-based \cite{Fu2014} & {\color{black}1.3531} &  {\color{black} 16.8757} & 0.6233\\
  UDCP  \cite{Drews2016} & 5.1300& 11.0296 &0.4999\\
  regression-based~\cite{Li2017prl} & 1.1365 & 17.5751 &0.6543 \\
  GDCP~\cite{Peng2018} & 3.6345 &12.5264 &0.5503\\
  Red Channel \cite{Galdran2015} &2.1073& 14.8935 &0.5973 \\
  histogram prior~\cite{Li2016} & 1.6282 & 16.0137& 0.5888\\
  blurriness-based~\cite{Peng2017} &1.5826 & 16.1371 & 0.6582\\
 dive+ &  \textcolor[rgb]{1.00,0.00,0.00}{0.5358} & \textcolor[rgb]{1.00,0.00,0.00}{20.8408} & \textcolor[rgb]{1.00,0.00,0.00}{0.8705}\\
  \hline
\end{tabular}
\vspace{\baselineskip}
\label{table_2}
\end{table}

\subsubsection{Non-reference Evaluation}

We employ two non-reference metrics (\ie, UCIQE \cite{Yang2015} and UIQM \cite{Panetta2016}) which are usually used for underwater image quality evaluation \cite{Li2017prl,Li2016,Peng2017,Peng2018}.
A higher UCIQE score indicates the result has better balance among the chroma, saturation, and contrast, while a higher UIQM score indicates the result is more consistent with human visual perception. The average scores are shown in Table~\ref{table_3}.

\begin{table}[htbp]
\caption{No-reference Image Quality Evaluation.}
 \centering
\begin{tabular}{c|c|c}
  \hline
  \textbf{Method}  & \textbf{UCIQE \cite{Yang2015}} $\uparrow$ &  \textbf{UIQM \cite{Panetta2016}} $\uparrow$\\
  \hline
  fusion-based \cite{Ancuti2012} &  \textcolor[rgb]{0.00,0.07,1.00}{0.6414} & 1.5310\\
  two-step-based \cite{Fu2017}  &0.5776 & 1.4002\\
  retinex-based \cite{Fu2014}  &0.6062 & 1.4338\\
  UDCP \cite{Drews2016}  &0.5852 & {\color{red}1.6297}\\
  regression-based~\cite{Li2017prl}  &0.5971 & 1.2996\\
  GDCP~\cite{Peng2018}  & 0.5993 & 1.4301\\
  Red Channel \cite{Galdran2015}  & 0.5421& 1.2147 \\
  histogram prior~\cite{Li2016}  & \textcolor[rgb]{1.00,0.00,0.00}{0.6778} & \textcolor[rgb]{0.00,0.07,1.00}{1.5440}\\
  blurriness-based~\cite{Peng2017}  &0.6001 &1.3757\\
  dive+ &   0.6227& 1.3410\\
  \hline
\end{tabular}
\vspace{\baselineskip}
\label{table_3}
\end{table}
In Table~\ref{table_3}, the histogram prior~\cite{Li2016} and UDCP \cite{Drews2016} obtain the highest scores of UCIQE and UIQM, respectively. The dive+ and fusion-based \cite{Ancuti2012} are no more the best performers. It is interesting that the good performers in terms of UCIQE and UIQM metrics are not consistent with the subjective pairwise comparisons, though both UCIQE and UIQM claim that they take the human visual perception into account. Such a result provides evidence that the current image quality evaluation metrics designed for underwater image are inconsistent with human visual perception in some cases. This is because humans have not evolved to see in aquatic habitats. When they are shown an underwater photo, they are most likely to pay attention to objects in the center of the scene, or whatever seems to be colorful or interesting (\eg, diver, fish, coral, \etc). Thus, human visual perception may be a totally inaccurate way of color correcting underwater images. It might be fine for visually pleasing images, but not to learn attenuation and backscatter.

\begin{figure}[!htp]
  \centering
\begin{minipage}[b]{0.32\linewidth}
  \centering
  \centerline{\includegraphics[width=2.5cm,height=2cm]{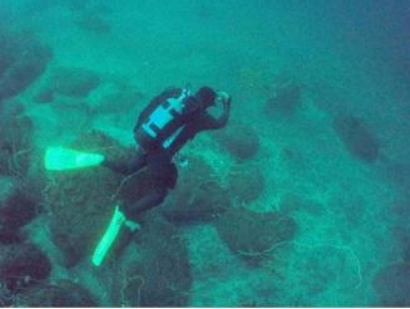}}
  \centerline{UCIQE/UIQM }\medskip
\end{minipage}
\begin{minipage}[b]{0.32\linewidth}
  \centering
  \centerline{\includegraphics[width=2.5cm,height=2cm]{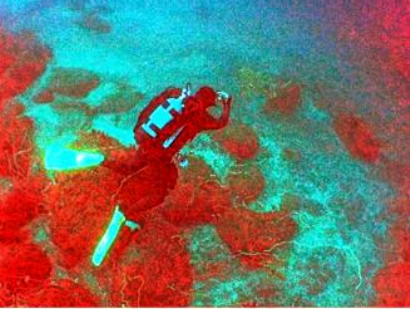}}
  \centerline{{\color{red}0.6941/1.7614}}\medskip
\end{minipage}
\begin{minipage}[b]{0.32\linewidth}
  \centering
  \centerline{\includegraphics[width=2.5cm,height=2cm]{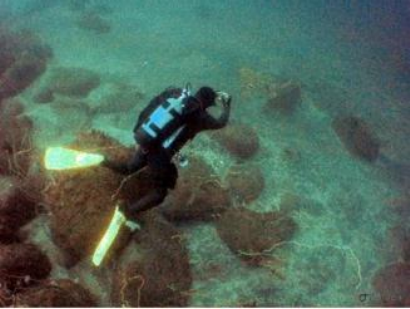}}
  \centerline{0.6139/1.4705}\medskip
\end{minipage}

\begin{minipage}[b]{0.32\linewidth}
  \centering
  \centerline{\includegraphics[width=2.5cm,height=2cm]{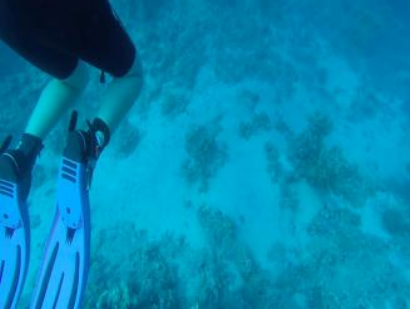}}
  \centerline{UCIQE/UIQM}\medskip
  \centerline{(a) Raws}\medskip
\end{minipage}
\begin{minipage}[b]{0.32\linewidth}
  \centering
  \centerline{\includegraphics[width=2.5cm,height=2cm]{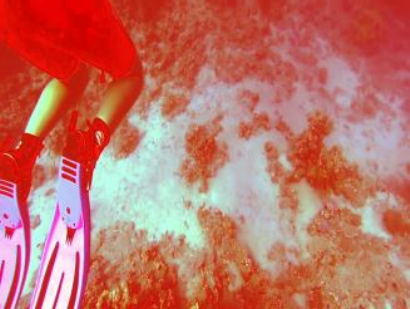}}
   \centerline{{\color{red}0.5945/1.0756}}\medskip
     \centerline{(b) histogram prior~\cite{Li2016}}\medskip
\end{minipage}
\begin{minipage}[b]{0.32\linewidth}
  \centering
  \centerline{\includegraphics[width=2.5cm,height=2cm]{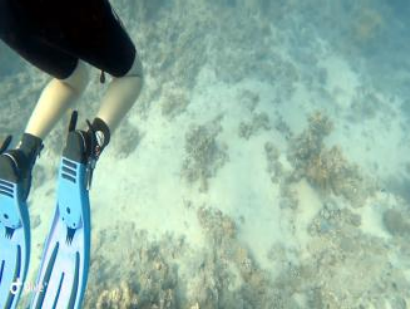}}
     \centerline{0.5708/0.9252}\medskip
  \centerline{(c) dive+}\medskip
\end{minipage}
\caption{Visual comparisons in terms of UCIQE and UIQM metrics. Higher scores are in red. It is obvious that higher quantitative scores do not lead to better subjectively quality.}
\label{fig:metric}
\vspace{-5mm}
\end{figure}

Furthermore, Figs.~\ref{fig_5}-\ref{fig_99} show the results generated by histogram prior~\cite{Li2016} and UDCP \cite{Drews2016} still suffer from color casts and over-enhancement. Through further analyzing, we found these two non-reference metrics  might be biased to some characteristics (not entire image) and did not take the color shift and artifacts into account. For example, the results with high contrast (\eg, the results of histogram prior~\cite{Li2016}) are usually favored by the UICQE metric.
To illustrate this phenomenon, we present an example in Fig.~\ref{fig:metric}.

In Fig.~\ref{fig:metric}, the results generated by histogram prior~\cite{Li2016} have obvious reddish color shift and artifacts; however, they obtain better quantitative scores in terms of UCIQE and UIQM metrics than the results of dive+. Thus, we believe there is a gap between the quantitative scores of non-reference metrics and the subjectively visual quality. In other words, the current image quality evaluation metrics designed for underwater image have limitations in some cases.

\subsubsection{Runtime Evaluation}

We compare the average runtime for the images of different sizes. Experiments are conducted by using MATLAB R2014b on a PC with an Intel(R) i7-6700 CPU, 32GB RAM. The average runtime is shown in Table~\ref{table_4}. The two-step-based \cite{Fu2017} is the fastest across different image sizes, while the retinex-based \cite{Fu2014} ranks the second fastest. The regression-based~\cite{Li2017prl} is the slowest method due to the time-consuming random forest-based transmission prediction, especially for images with large sizes.

\begin{table}[htbp]
\caption{Average Runtime for Different Image Sizes (in second).}
 \centering
\begin{tabular}{c|c|c|c}
  \hline
  \textbf{Method} &  \textbf{500 $\times$ 500} & \textbf{640 $\times$ 480} &  \textbf{1280 $\times$ 720}\\
  \hline
  fusion-based \cite{Ancuti2012} &  \textcolor[rgb]{0.00,0.07,1.00}{0.6044} & \textcolor[rgb]{0.00,0.07,1.00}{0.6798} & \textcolor[rgb]{0.00,0.07,1.00}{1.8431}\\
  two-step-based \cite{Fu2017} &\textcolor[rgb]{1.00,0.00,0.00}{0.2978}  & \textcolor[rgb]{1.00,0.00,0.00}{0.4391}& \textcolor[rgb]{1.00,0.00,0.00}{1.0361}\\
  retinex-based \cite{Fu2014}  &  0.6975 & 0.8829 & 2.1089\\
  UDCP \cite{Drews2016} &  2.2688 & 3.3185 & 9.9019\\
  regression-based~\cite{Li2017prl} & 138.6138 & 167.1711 & 415.4935\\
  GDCP~\cite{Peng2018}  & 3.2676& 3.8974& 9.5934\\
  Red Channel  \cite{Galdran2015}  & 2.7523& 3.2503 & 9.7447\\
  histogram prior~\cite{Li2016}  & 4.6284 &5.8289 &16.9229\\
  blurriness-based~\cite{Peng2017}   & 37.0018& 47.2538& 146.0233\\
  \hline
\end{tabular}
\vspace{\baselineskip}
\label{table_4}
\end{table}

\begin{figure*}[!htb]
\centering
\includegraphics[width=15cm,height=8cm]{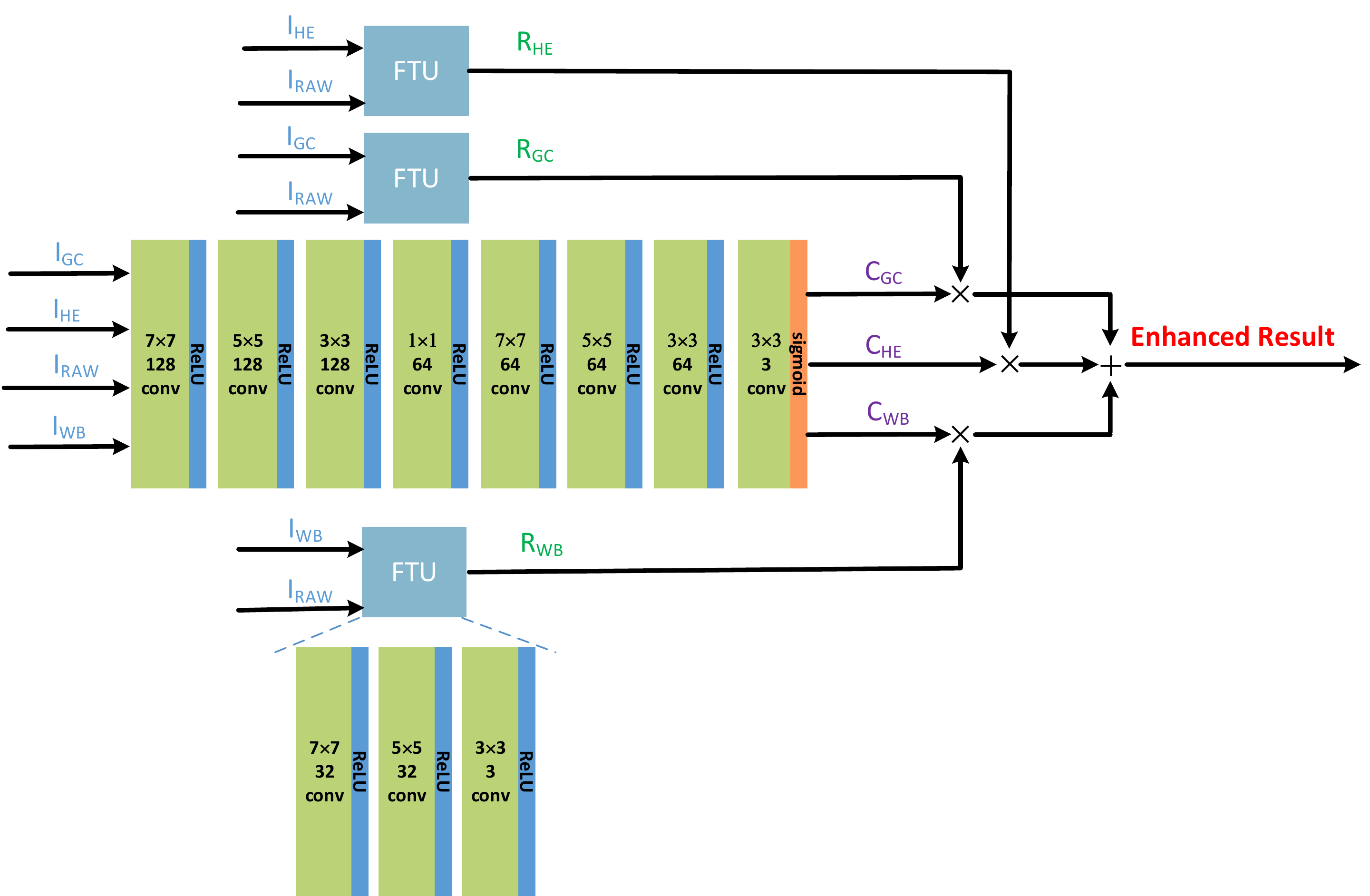}
\caption{An overview of the proposed Water-Net architecture. Water-Net is a gated fusion network, which fuses the inputs with the predicted confidence maps to achieve the enhanced result. The inputs are first transferred to the refined inputs by the Feature Transformation Units (FTUs) and then the confidence maps are predicted. At last, the enhanced result is achieved by fusing the refined inputs and the corresponding confidence maps.}
\label{fig_9}
\end{figure*}

After reviewing and evaluating the state-of-the-art underwater image enhancement methods, we found that the fusion-based \cite{Ancuti2012} is the relatively best performer in most cases, while other compared methods have obvious disadvantages. However, there is no method which always wins when facing a large-scale real-world underwater image dataset (\ie, UIEB). All in all, due to neglecting the underwater imaging physical models, the non-physical model-based methods, such as two-step-based \cite{Fu2017} and retinex-based \cite{Fu2014}, produce over-/under-enhanced results. Physical model-based methods, such as UDCP \cite{Drews2016}, employ an outdoor haze formation based model to predict the medium transmission which is not well-suited for the underwater scenario. Inaccurate physical models and assumptions result in color casts and remaining haze in the results such as regression-based~\cite{Li2017prl}, GDCP~\cite{Peng2018}, Red Channel \cite{Galdran2015}, histogram prior~\cite{Li2016}, and blurriness-based~\cite{Peng2017}. Some methods, such as retinex-based \cite{Fu2014} and histogram prior~\cite{Li2016}, tend to introduce noise and artifacts, which leads to visually unpleasing results. The runtime of some methods seriously limits their practical applications.

In the future, a comprehensive method that can robustly deal with a variety of underwater image degradation is expected. The non-reference metrics which are more effective and consistent with human visual perception are desired in the community of underwater image enhancement. More discussions will be provided in Sec. VI.

\section{Proposed Model}
Despite the remarkable progress of underwater image enhancement methods, the generalization of deep learning-based underwater image enhancement models still falls behind the conventional state-of-the-art methods due to the lack of effective training data and well-designed network architectures. With the UIEB, we propose a CNN model for underwater image enhancement, called Water-Net.
The purpose of the proposed Water-Net as a baseline is to call for the development of deep learning-based underwater image enhancement, and demonstrate the generalization of the UIEB for training CNNs. Note that the proposed Water-Net is only a baseline model which can be further improved by well-designed network architectures, task-related loss functions, and the like.

In this section, we first present input generation from an underwater image and the architecture of the proposed Water-Net. Then we present the training and implementation details. At last, we perform experiments to demonstrate its advantages.

\subsection{Input Generation}
As discussed in Sec. IV, there is no algorithm generalized to all types of underwater images due to the complicated underwater environment and lighting conditions. In general, the fusion-based \cite{Ancuti2012} achieves decent results, which benefits from the inputs derived by multiple pre-processing operations and a fusion strategy.
In the proposed Water-Net, we also employ such a manner. Based on the characteristics of underwater image degradation, we generate three inputs by respectively applying White Balance (WB), Histogram Equalization (HE) and Gamma Correction (GC) algorithms to an underwater image. Specifically, WB algorithm is used to correct the color casts, while HE and GC algorithms aim to improve the contrast and lighten up dark regions, respectively. We directly employ the WB algorithm proposed in \cite{Ancuti2012}, whose effectiveness has been turned out. For the HE algorithm, we apply the \emph{adapthisteq} function \cite{CLAHE} provided by MATLAB to the L component in Lab color space, and then transform back into RGB color space. We set the Gamma value of GC algorithm to 0.7 empirically.

\subsection{Network Architecture}
Water-Net employs a gated fusion network architecture to learn three confidence maps which will be used to combine the three input images into an enhanced result. The learned confidence maps determine the most significant features of inputs remaining in the final result. The impressive performance of the fusion-based underwater image enhancement method \cite{Ancuti2012} also encourages us to explore the fusion-based networks.

The architecture of the proposed Water-Net and parameter settings are shown in Fig.~\ref{fig_9}. As a baseline model, the Water-Net is a plain fully CNN. We believe that the widely used backbones such as the U-Net architecture \cite{Unet} and the residual network architecture \cite{Residual} can be incorporated to improve the performance. We feed the three derived inputs and original input to the Water-Net to predict the confidence maps. Before performing fusion, we add three Feature Transformation Units (FTUs) to refine the three inputs. The purpose of the FTU is to reduce the color casts and artifacts introduced by the WB, HE, and GC algorithms.  At last, the refined three inputs are multiplied by the three learned confidence maps to achieve the final enhanced result:
\begin{equation}
\label{equ_1}
I_{en}=R_{WB}\odot C_{WB}+R_{HE}\odot C_{HE}+R_{GC}\odot C_{GC},
\end{equation}
where $I_{en}$ is the enhanced result; $\odot$ indicates the element-wise production of matrices; $R_{WB}$, $R_{HE}$, and $R_{GC}$ are the refined results of input after processing by WB, HE, and GC algorithms, respectively; $C_{WB}$, $C_{HE}$, and $C_{GC}$ are the learned confidence maps.

\subsection{Implementations}

A random set of 800 pairs of the images extracted from the UIEB is used to generate the training set. We resize the training data to size 112$\times$112 due to our limited memory. Flipping and rotation are used to obtain 7 augmented versions of original training data.
Resizing the training data to a fixed size is widely used in deep learning, such as image dehazing \cite{Ren2016}, salient object detection \cite{Saliency}, \etc.
In contrast to image super-resolution (\eg, \cite{SRCNN}) that has the same degradation in different regions of an input image, different regions of an underwater image may have different degradation. Thus, the contextual information of an underwater image is important for network optimization. This is the main reason why we do not use image patches to train our network.
The rest 90 pairs of the images in the UIEB are treated as the testing set.

To reduce the artifacts induced by pixel-wise loss functions such as $\ell_{1}$ and $\ell_{2}$, we minimize the perceptual loss function to learn the mapping function of underwater image enhancement. The perceptual loss can produce visually pleasing and realistic results, which has been widely used in image restoration and synthesis networks, such as image super-resolution \cite{Johnson}, photographic image synthesis \cite{Chen2017}, \etc.
		
Inspired by \cite{Johnson}, we define the perceptual loss based on the ReLU activation layers (\ie, layer relu5\_4) of the pre-trained 19 layers VGG network \cite{Simonyan}. Let $\phi_{j}(x)$ be the $j$th convolution layer (after activation) of the VGG19 network $\phi$ pretrained on the ImageNet dataset \cite{Deng}. The perceptual loss is expressed as the distance between the feature representations of the enhanced image $I_{en}$ and the reference image $I_{gt}$:
\begin{equation}
\label{equ_2}
L^{\phi}_{j}=\frac{1}{C_{j}H_{j}W_{j}}\sum_{i=1}^{N}\|\ \phi_{j}(I_{en}^{i})-\phi_{j}(I_{gt}^{i})\|,
\end{equation}
where $N$ is the number of each batch in the training procedure; $C_{j}H_{j}W_{j}$ represents the dimension of the feature maps of the $j$th convolution layer within the VGG19 network. $C_{j}$, $H_{j}$, and $W_{j}$ are the number, height, and width of the feature map.

We implemented the proposed Water-Net with TensorFlow on a PC with an Nvidia $1080$Ti GPU. During training, a batch-mode learning method with a batch size of $16$ was applied. The filter weights of each layer were initialized by standard Gaussian distribution. Bias was initialized as a constant. We used ADAM with default parameters for our network optimization. We initialized the learning rate to $1e^{-3}$ and decreased the learning rate by 0.1 every 10,000 iterations until the Water-Net converges. Our Water-Net can process an image with a size of 640 $\times$ 480 within 0.128s (8FPS).

\subsection{Experiments}

\begin{figure*}[!htb]
\centering
\includegraphics[width=18cm,height=4.8cm]{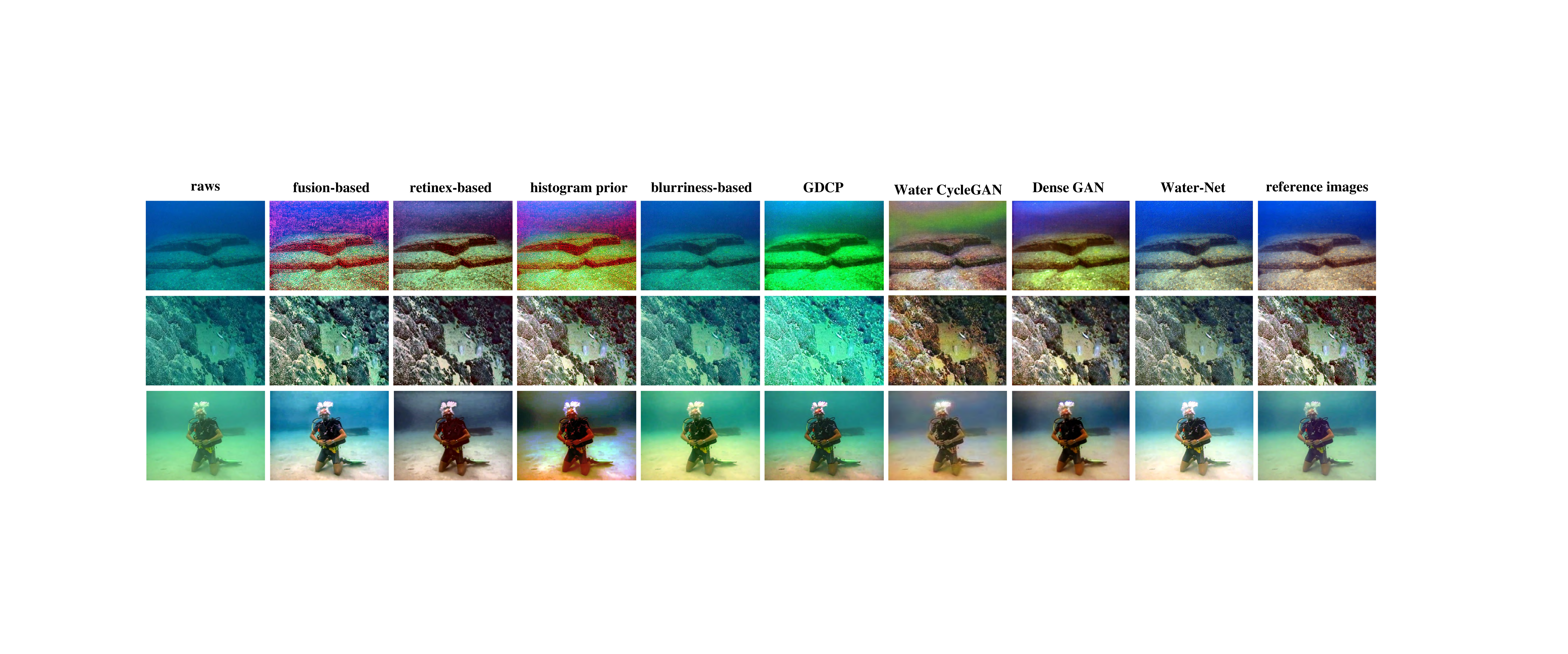}
\caption{Subjective comparisons on underwater images from testing set. From left to right are raw underwater images, and the results of fusion-based \cite{Ancuti2012}, retinex-based \cite{Fu2014}, histogram prior~\cite{Li2016}, blurriness-based~\cite{Peng2017}, GDCP~\cite{Peng2018}, Water CycleGAN \cite{Emerging}, Dense GAN \cite{Guo2019}, the proposed Water-Net, and reference images.}
\label{fig_10}
\end{figure*}

\begin{figure*}[!htb]
\centering
\includegraphics[width=18cm,height=5cm]{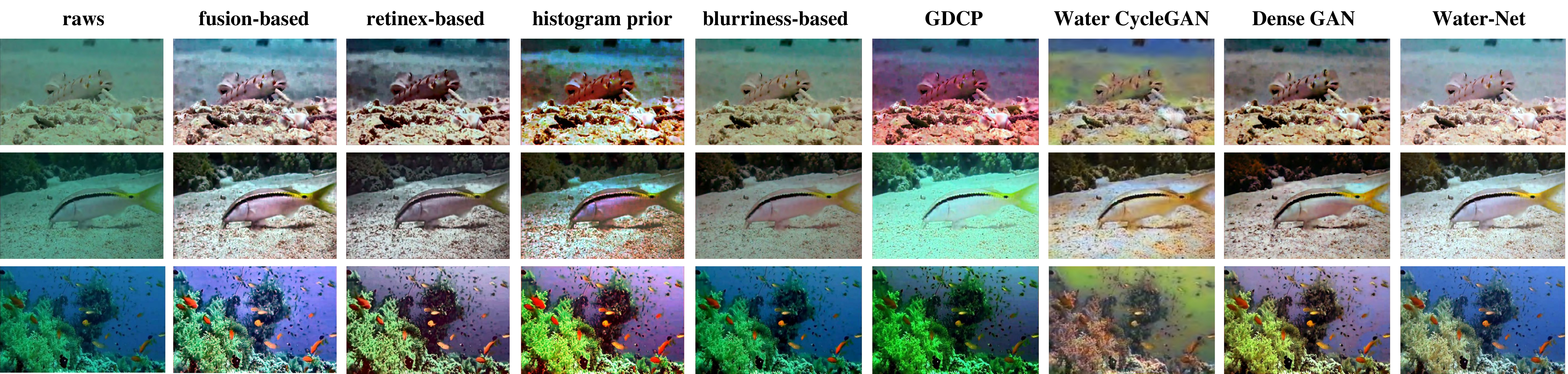}
\caption{Subjective comparisons on underwater images from challenging set. From left to right are raw underwater images, and the results of fusion-based \cite{Ancuti2012}, retinex-based \cite{Fu2014}, histogram prior~\cite{Li2016}, blurriness-based~\cite{Peng2017}, GDCP~\cite{Peng2018}, Water CycleGAN \cite{Emerging}, Dense GAN \cite{Guo2019}, and the proposed Water-Net.}
\label{fig_11}
\end{figure*}

To demonstrate the advantages achieved by the proposed Water-Net, we compare it against several state-of-the-art underwater image enhancement methods. The experiments are conducted on the testing set which includes 90 underwater images and the challenging set including 60 underwater images. We show several results in Figs.~\ref{fig_10} and \ref{fig_11}.

In Fig.~\ref{fig_10}, the proposed Water-Net effectively removes the haze on the underwater images and remits color casts, while the competing  methods introduce unexpected colors (\eg, fusion-based \cite{Ancuti2012}, GDCP~\cite{Peng2018}, histogram prior~\cite{Li2016}, Water CycleGAN \cite{Emerging}, and Dense GAN \cite{Guo2019}) and artifacts (\eg, fusion-based \cite{Ancuti2012}, retinex-based \cite{Fu2014}, histogram prior~\cite{Li2016}, Water CycleGAN \cite{Emerging}, and Dense GAN \cite{Guo2019}) or have little effect on inputs (\eg, blurriness-based~\cite{Peng2017}). In addition, it is interesting that our results even achieve better visual quality than the corresponding reference images (\eg, more natural appearance and better details). This is because that the perceptual loss optimized Water-Net can learn the potential attributes of good visual quality from the large-scale real-world underwater image dataset. For the results on challenging set shown in Fig.~\ref{fig_11}, the proposed Water-Net produces visually pleasing results. By contrast, other methods tend to introduce artifacts, over-enhancement (\eg, foregrounds), and color casts (\eg, reddish or greenish color).

\begin{table}[htbp]
\caption{Full-reference Image Quality Assessment in Terms of MSE, PSNR, and SSIM on Testing Set.}
 \centering
\begin{tabular}{c|c|c|c}
  \hline
  \textbf{Method} & \textbf{MSE ($\times 10^3$)} $\downarrow$& \textbf{PSNR (dB)} $\uparrow$ & \textbf{SSIM} $\uparrow$ \\
  \hline
  fusion-based \cite{Ancuti2012} & \textcolor[rgb]{0.00,0.07,1.00}{1.1280} & \textcolor[rgb]{0.00,0.07,1.00}{17.6077} & \textcolor[rgb]{0.00,0.07,1.00}{0.7721}\\
  retinex-based \cite{Fu2014} & 1.2924 & 17.0168 & 0.6071\\
  GDCP~\cite{Peng2018} & 4.0160 & 12.0929&0.5121\\
  histogram prior~\cite{Li2016} & 1.7019 & 15.8215
  & 0.5396\\
  blurriness-based~\cite{Peng2017} & 1.9111& 15.3180&0.6029 \\
  Water CycleGAN \cite{Emerging} &  1.7298 & 15.7508 & 0.5210\\
  Dense GAN \cite{Guo2019} &  1.2152 &  17.2843 & 0.4426 \\
  Water-Net &  \textcolor[rgb]{1.00,0.00,0.00}{0.7976} & \textcolor[rgb]{1.00,0.00,0.00}{19.1130} & \textcolor[rgb]{1.00,0.00,0.00}{0.7971}\\
  \hline
\end{tabular}
\vspace{\baselineskip}
\label{table_5}
\end{table}

Table~\ref{table_5} reports the quantitative results of different methods in terms of MSE, PSNR, and SSIM on the testing set. The quantitative results are obtained by comparing the result of each method with the corresponding reference image.  We discard the non-reference metrics designed for underwater image enhancement based on the conclusion drawn in Sec. IV. Our Water-Net achieves the best performance in terms of full-reference image quality assessment.
In addition, instead of pairwise comparison, we conduct a user study to score the visual quality of the results on challenging set. This is because some images in the challenging set are too difficult to obtain satisfactory results following the procedure of reference image generation. Thus, we invited 50 participants (the same volunteers with the reference image generation) to score results.  The scores have five scales ranging from 5 to 1 which represent ``Excellent'', ``Good'', ``Fair'', ``Poor'' and ``Bad'', respectively. The average scores of the results by each method on challenging set are shown in Table~\ref{table_6}. Besides, we also provide the standard deviation of the results by each method on challenging set. We exclude the scores of Water CycleGAN \cite{Emerging} and Dense GAN \cite{Guo2019} due to their obviously unpleasing results as shown in Fig.~\ref{fig_11}. Our Water-Net receives the highest average score and lowest standard deviation, which indicates our method produces better results from a subjective perspective and has more robust performance.

\begin{table}[htbp]
\caption{The Average Scores and Standard Deviation of Different Methods on Challenging Set.}
 \centering
\begin{tabular}{c|c|c}
  \hline
  \textbf{Method} & \textbf{Average Score} $\uparrow$ & \textbf{Standard Deviation} $\downarrow$\\
  \hline
  fusion-based \cite{Ancuti2012} & \textcolor[rgb]{0.00,0.07,1.00}{2.28} & 0.8475\\
  retinex-based \cite{Fu2014} & 2.23 & 0.8720 \\
  GDCP~\cite{Peng2018} & 1.90 & 0.8099\\
  histogram prior~\cite{Li2016} & 2.08 &  0.7897\\
  blurriness-based~\cite{Peng2017} & 2.02 &\textcolor[rgb]{0.00,0.07,1.00}{0.7762}\\
  Water-Net &  \textcolor[rgb]{1.00,0.00,0.00}{2.57} & \textcolor[rgb]{1.00,0.00,0.00}{0.7280} \\
  \hline
\end{tabular}
\vspace{\baselineskip}
\label{table_6}
\end{table}

Qualitative and quantitative experiments demonstrate the effectiveness of the proposed Water-Net and also indicate the constructed dataset can be used for training CNNs. However, there is room for the improvement of underwater image enhancement. Besides, underwater images in the challenging set still cannot be enhanced well.

\section{Conclusion, Limitations, and Future Work}

In this paper, we have constructed an underwater image enhancement benchmark dataset which offers large-scale real underwater images and the corresponding reference images. This benchmark dataset enables us to comprehensively study the existing underwater image enhancement methods, and easily train CNNs for underwater image enhancement.
As analyzed in qualitative and quantitative evaluations, there is no method which always wins in terms of full- and no-reference metrics.
%
In addition, effective non-reference underwater image quality evaluation metrics are highly desirable. To promote the development of deep learning-based underwater image enhancement methods, we proposed an underwater image enhancement CNN trained by the constructed dataset. Experimental results demonstrate the proposed CNN model performs favorably against the state-of-the-art methods, and also verify the generalization of the constructed dataset for training CNNs.

Although our reference image generation strategy can select visually pleasing results, there is a problem that affects the selection of reference images and further limits the performance of our network. Specifically, the effect of backscatter is difficult to be completely removed, especially for the backscatter in far distances. We use the state-of-the-art image enhancement algorithms to process the raw underwater images; however, the backscatter still cannot be completely removed in some cases. Despite our constructed dataset is not dominated by such underwater images, the backscatter is always discouraged in the reference images. Moreover, some invited volunteers cannot identify the exponentially increasing effect of backscatter in scenes with large ranges. In summary, the main reasons for the shortcomings of our reference image generation  are: 1) the existing algorithms follow inaccurate image formation models or assumptions which inherently limit the performance of underwater image enhancement; and 2) some volunteers do not understand the underwater imaging physical model, thus they may ignore the effect of the presence of backscatter in far ranges.  Note that the use of inaccurate imaging models is a major problem which keeps the field of underwater computer vision at standstill. Fortunately, the recent work \cite{Akkaynak2019} sheds light on the future research of underwater vision.

In future work, we will extend the constructed dataset towards more challenging underwater images and underwater videos. Moreover, we will try to design a range map estimation network. The provided 1100 underwater images with range maps in \cite{Akkaynak2019} could be used for the range map estimation network training. With the estimated range maps, we will make full use of such key prior information to further improve the performance of underwater image enhancement network. Besides, inspired by recent work \cite{Akkaynak2017,Akkaynak2019}, we believe that more physically reasonable underwater image enhancement algorithms will arise. At that time, we will re-organize the selection of the reference images from more reliable results and also further train the volunteers on what the degrading effects of attenuation and backscatter are, and what it looks like when either is improperly corrected.
Additionally, the main purpose of constructing the real-world underwater dataset in this paper is to evaluate the state-of-the-art underwater image enhancement methods and provide paired training data for deep models. Since the full-reference metrics and training a deep model only need a single reference, we do not select multiple references or define the image quality level. However, the image quality level of multiple reference images does help in underwater image enhancement. Thus, we will provide multiple reference images for an underwater image and define the image quality level of their reference images when we re-organize the selection of the reference images.



\bibliographystyle{IEEEtran.bst}

\begin{IEEEbiography}[{\includegraphics[width=1in,height=1.25in,clip,keepaspectratio]{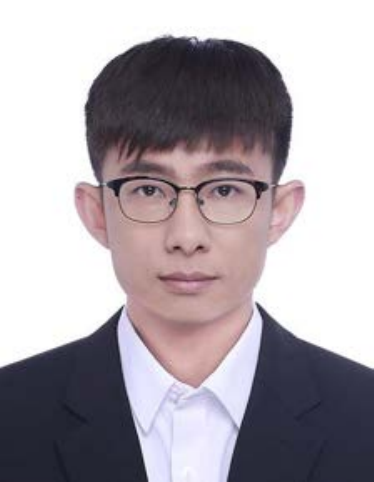}}]{Chongyi Li}
received his Ph.D. degree with the School of Electrical and Information Engineering, Tianjin University, Tianjin, China in June 2018. During 2016 to 2017, he was a joint-training Ph.D student in Australian National University, Australia.\par
He is currently a Postdoc Research Fellow at the Department of Computer Science, City University of Hong Kong (CityU), Hong Kong SAR, China. His current research focuses on image processing, computer vision, and deep learning, particularly in the domains of image restoration and enhancement. He is a Guest Editor of special issues for the Signal Processing: Image Communication.
\end{IEEEbiography}

\begin{IEEEbiography}[{\includegraphics[width=1in,height=1.25in,clip,keepaspectratio]{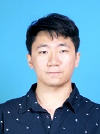}}]{Chunle Guo}
received his B.S. degree from School of Electronic Information Engineering in Tianjin University. He is currently pursuing his Ph.D. degree with the School of Electrical and Information Engineering, Tianjin University, Tianjin, China. His current research focuses on image processing and computer vision, particularly in the domains of deep learning-based image restoration and enhancement.
\end{IEEEbiography}

\begin{IEEEbiography}[{\includegraphics[width=1in,height=1.25in,clip,keepaspectratio]{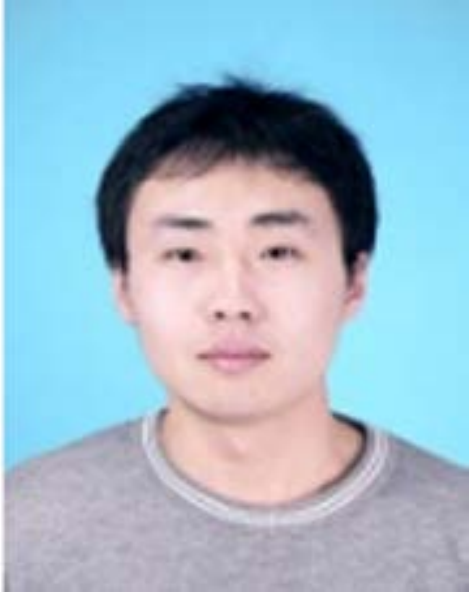}}]{Wenqi Ren}
is an assistant professor in Institute of Information Engineering, Chinese Academy of Sciences, China. He received his Ph.D. degree from Tianjin University in 2017. During 2015 to 2016, he was a joint-training Ph.D. student in Electrical Engineering and Computer Science at the University of California, Merced, CA, USA. His research interest includes image/video analysis and enhancement, and related vision problems.
\end{IEEEbiography}

\vspace{-10 mm}
\begin{IEEEbiography}[{\includegraphics[width=1in,height=1.25in,clip,keepaspectratio]{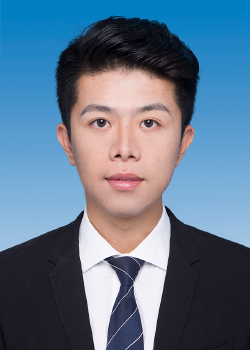}}]{Runmin Cong}(M'19)
received the Ph.D. degree in information and communication engineering from Tianjin University, Tianjin, China, in June 2019.
He is currently an Associate Professor with the Institute of Information Science, Beijing Jiaotong University, Beijing, China. He was a visiting student at Nanyang Technological University (NTU), Singapore, from Dec. 2016 to Feb. 2017. Since May 2018, he has spent one year as a Research Associate at the Department of Computer Science, City University of Hong Kong (CityU), Hong Kong. His research interests include computer vision and intelligent video analysis, multimedia information processing, saliency detection and segmentation, remote sensing image interpretation, and deep learning.\par
He is a Reviewer for the IEEE TIP, TMM, and TCSVT, etc, and is a Guest Editor of special issues for the Signal Processing: Image Communication. Dr. Cong was a recipient of the Best Student Paper Runner-Up at IEEE ICME in 2018, the First Prize for Scientific and Technological Progress Award of Tianjin Municipality, and the Excellent Doctoral Degree Dissertation Award from BSIG.
\end{IEEEbiography}
\vspace{-10 mm}
\begin{IEEEbiography}[{\includegraphics[width=1in,height=1.25in,clip,keepaspectratio]{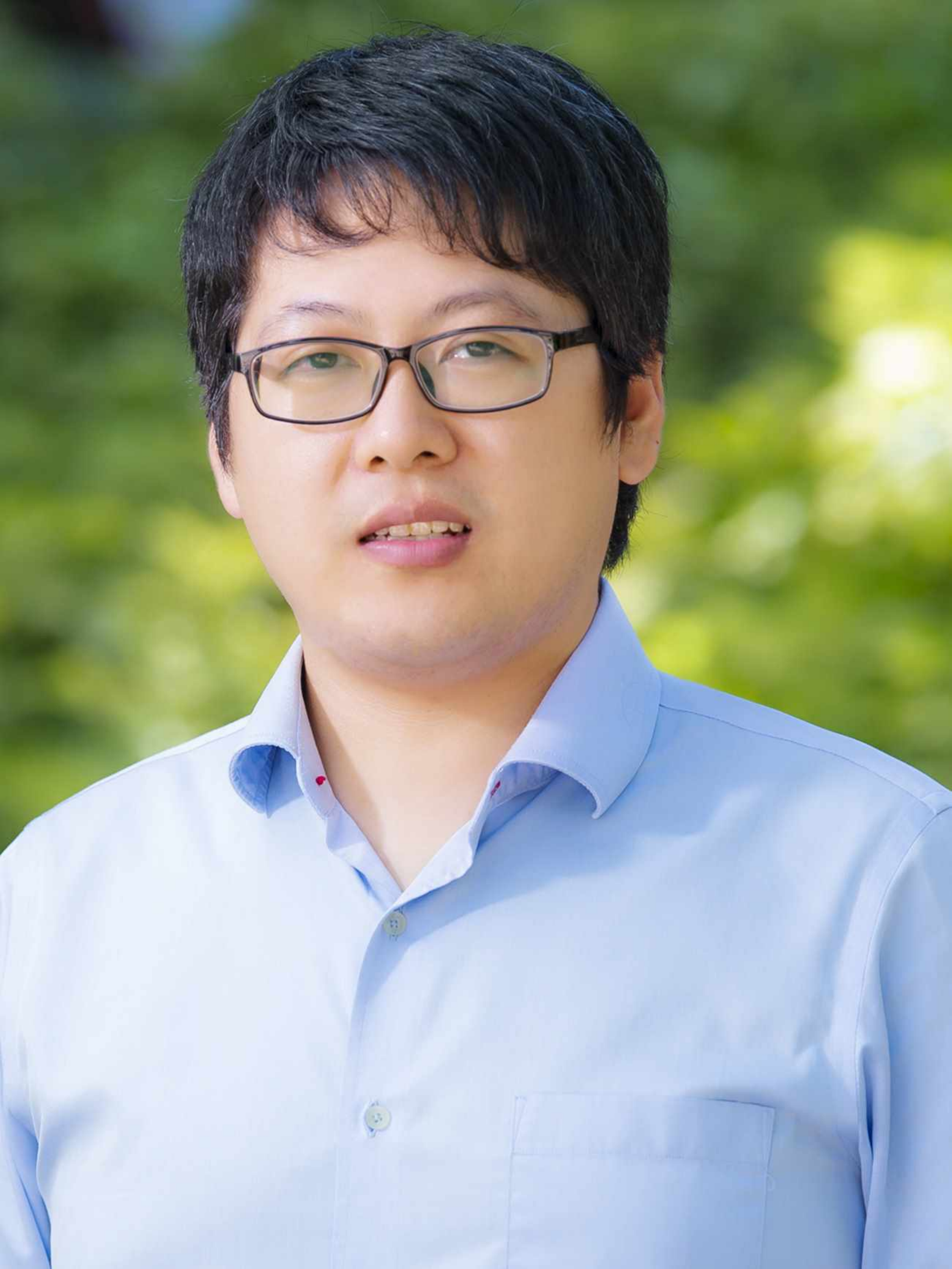}}]{Junhui Hou}
(S'13-M'16) received the B.Eng. degree in information engineering (Talented Students Program) from the South China University of Technology, Guangzhou, China, in 2009, the M.Eng. degree in signal and information processing from Northwestern Polytechnical University, Xian, China, in 2012, and the Ph.D. degree in electrical and electronic engineering from the School of Electrical and Electronic Engineering, Nanyang Technological University, Singapore, in 2016. He has been an Assistant Professor with the Department of Computer Science, City University of Hong Kong, since 2017. His research interests fall into the general areas of visual signal processing, such as image/video/3D geometry data representation, processing and analysis, semi-supervised/unsupervised data modeling for clustering/classification, and data compression and adaptive transmission.\par
Dr. Hou was the recipient of several prestigious awards, including the Chinese Government Award for Outstanding Students Study Abroad from China Scholarship Council in 2015, and the Early Career Award from the Hong Kong Research Grants Council in 2018. He is serving/served as an Associate Editor for The Visual Computer, an Area Editor for Signal Processing: Image Communication, the Guest Editor for the IEEE Journal of Selected Topics in Applied Earth Observations and Remote Sensing, the Journal of Visual Communication and Image Representation, and Signal Processing: Image Communication, and an Area Chair of ACM International Conference on Multimedia (ACM MM) 2019 and IEEE International Conference on Multimedia \& Expo (IEEE ICME) 2020.
\end{IEEEbiography}
\vspace{-10 mm}
\begin{IEEEbiography}[{\includegraphics[width=1in,height=1.25in,clip,keepaspectratio]{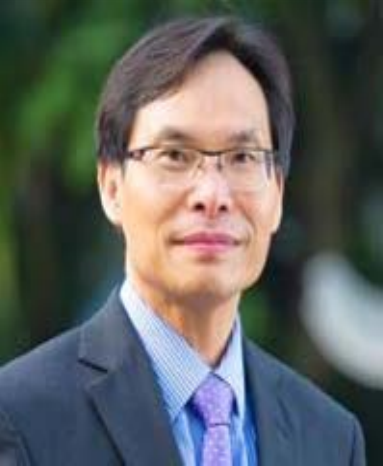}}]{Sam Kwong}
(M'93-SM'04-F'13) received the B.S. degree in electrical engineering from The State University of New York at Buffalo, Buffalo, NY, USA, in 1983, the M.S. degree from the University of Waterloo, Waterloo, ON, Canada, in 1985, and the Ph.D. degree from the University of Hagen, Germany, in 1996. From 1985 to 1987, he was a Diagnostic Engineer with Control Data Canada, Mississauga, ON, Canada. He was a Member of Scientific Staff with Bell Northern Research Canada, Ottawa. In 1990, he became a Lecturer with the Department of Electronic Engineering, City University of Hong Kong, where he is currently a Chair Professor with the Department of Computer Science. His research interests are video and image coding and evolutionary algorithms.
\end{IEEEbiography}
\vspace{-10 mm}
\begin{IEEEbiography}[{\includegraphics[width=1in,height=1.25in,clip,keepaspectratio]{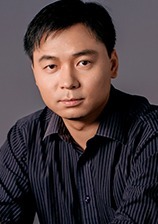}}]{Dacheng Tao}
(F'15) is Professor of Computer Science and ARC Laureate Fellow in the School of Computer Science and the Faculty of Engineering, and the Inaugural Director of the UBTECH Sydney Artificial Intelligence Centre, at The University of Sydney. His research results in artificial intelligence have expounded in one monograph and 200+ publications at prestigious journals and prominent conferences, such as IEEE T-PAMI, IJCV, JMLR, AAAI, IJCAI, NIPS, ICML, CVPR, ICCV, ECCV, ICDM, and KDD, with several best paper awards. He received the 2018 IEEE ICDM Research Contributions Award and the 2015 Australian Scopus-Eureka prize. He is a Fellow of the IEEE and the Australian Academy of Science.
\end{IEEEbiography}

\flushbottom

\end{document}